\documentclass[10pt,twocolumn,letterpaper]{article}

\usepackage{iccv}
\usepackage{times}
\usepackage{epsfig}
\usepackage{graphicx}
\usepackage{amsmath}
\usepackage{amssymb}

\usepackage{graphbox}
\usepackage{amsmath}
\usepackage{amssymb}
\usepackage{booktabs}
\usepackage{svg}
\usepackage{xfrac}
\newcommand*\circled[1]{\tikz[baseline=(char.base)]{
            \node[shape=circle,draw,inner sep=.6pt] (char) {#1};}}
\usepackage[ruled,vlined]{algorithm2e}
\usepackage{caption,subcaption}
\usepackage[font=small]{caption}
\usepackage{tikz}
\usepackage{paralist} % Inline lists
\usepackage{comment}
\usepackage{amsmath,amssymb} % define this before the line numbering.e
\usepackage{xcolor,colortbl}
\definecolor{shadecolor}{RGB}{255,203,203}
\definecolor{lightgreen}{RGB}{39,179,118}
\definecolor{student}{RGB}{76,115,176}
\definecolor{baseline}{RGB}{221,133,83}
\definecolor{teacher}{RGB}{85,168,105}
\usepackage{paralist} % Inline lists
\usepackage{cite}
\usepackage{wrapfig}
\usepackage[accsupp]{axessibility}  % Improves PDF readability for those with disabilities.
\newcommand{\NA}{---}
\definecolor{redred}{RGB}{196,23,63}
\definecolor{pinkpink}{RGB}{243,157,153}
\definecolor{greengreen}{RGB}{200,228,206}
\definecolor{blueblue}{RGB}{168,218,224}
\definecolor{orangeorange}{RGB}{231,96,83}
\definecolor{graygray}{RGB}{229,229,229}
\definecolor{yellowyellow}{RGB}{255,229,153}
\SetAlFnt{\tiny}
\SetAlCapFnt{\small}
\SetAlCapNameFnt{\small}
\definecolor{commentcolor}{RGB}{110,154,155}   % define comment color
\usepackage{tikz}
\usepackage{colortbl}

\makeatletter
\newcommand{\printfnsymbol}[1]{%
  \textsuperscript{\@fnsymbol{#1}}%
}
\makeatother

\usepackage[pagebackref=true,breaklinks=true,letterpaper=true,colorlinks,bookmarks=false]{hyperref}

\iccvfinalcopy % *** Uncomment this line for the final submission

\def\iccvPaperID{12277} % *** Enter the ICCV Paper ID here

\ificcvfinal\fi

\begin{document}

%%%%%%%%% TITLE
\title{Multimodal Distillation for Egocentric Action Recognition}

% \author{
% Tim Lebailly$^{1*}$ \hspace{1cm} Thomas Stegm\"uller$^{2*}$ \hspace{1cm} Behzad Bozorgtabar$^{2,3}$ \\
% Jean-Philippe Thiran$^{2,3}$ \hspace{1cm} Tinne Tuytelaars$^{1}$ \\ 
% $^{1}$KU Leuven \hspace{1cm} $^{2}$EPFL \hspace{1cm} $^{3}$CHUV \\ 
% {\small $^{1}$\texttt{\{firstname\}.\{lastname\}@esat.kuleuven.be}
% }

\author{Gorjan Radevski\thanks{\ \ Authors contributed equally.} \hspace{1cm}
% KU Leuven\\
\and
Dusan Grujicic\printfnsymbol{1} \hspace{1cm}
% KU Leuven\\
\and
Marie-Francine Moens \hspace{1cm}
\and
Matthew Blaschko \hspace{1cm}
\and
Tinne Tuytelaars \hspace{1.9cm}
\and
KU Leuven University, Belgium\\
{\tt\small \{firstname\}.\{lastname\}@kuleuven.be}
}

\maketitle
\ificcvfinal\thispagestyle{empty}\fi

%%%%%%%%% ABSTRACT
\begin{abstract}
The focal point of egocentric video understanding is modelling hand-object interactions. Standard models, e.g.\ CNNs or Vision Transformers, which receive RGB frames as input perform well, however, their performance improves further by employing \textit{additional} input modalities that provide complementary cues, such as object detections, optical flow, audio, etc. The added complexity of the modality-specific modules, on the other hand, makes these models impractical for deployment. The goal of this work is to retain the performance of such a multimodal approach, while using \textit{only} the RGB frames as input at inference time. We demonstrate that for egocentric action recognition on the Epic-Kitchens and the Something-Something datasets, students which are taught by multimodal teachers tend to be more accurate and better calibrated than architecturally equivalent models trained on ground truth labels in a unimodal or multimodal fashion. We further adopt a principled multimodal knowledge distillation framework, allowing us to deal with issues which occur when applying multimodal knowledge distillation in a na\"ive manner. Lastly, we demonstrate the achieved reduction in computational complexity, and show that our approach maintains higher performance with the reduction of the number of input views. We release our code at: \href{https://github.com/gorjanradevski/multimodal-distillation}{https://github.com/gorjanradevski/multimodal-distillation}
\end{abstract}

\section{Introduction}\label{sec:introduction}

The purpose of egocentric vision is enabling machines to interpret real-world data taken from a human's perspective. Its applications are numerous, ranging from recognizing \cite{zhou2015temporal} or anticipating \cite{furnari2019would} actions, to more complex tasks such as recognizing egocentric object-state changes, localizing action instances of a particular video moment \cite{grauman2022ego4d}, etc. The focal point of egocentric vision is hand-object interactions. Usually, these hand-object interactions take place in cluttered environments, where the object of interest is often occluded, or occurs only during a short time period. Furthermore, egocentric vision often suffers from motion blur -- due to the movement of the scene objects or the camera itself -- and thus, understanding video content from RGB frames alone may be challenging.\par

\begin{figure}[t]
\centering
\includegraphics[width=\columnwidth]{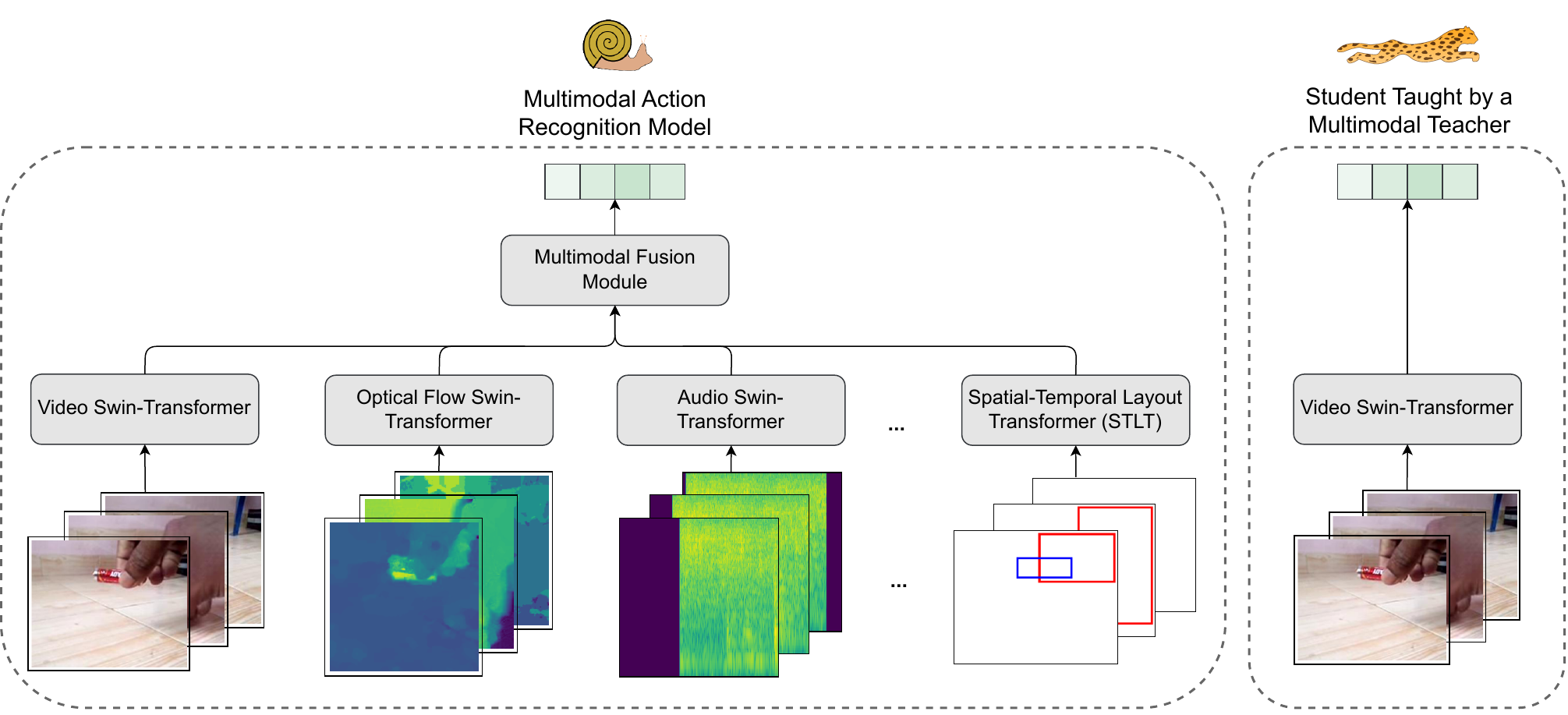}
\caption{\textbf{Motivation:} The multimodal action recognition model is powerful, but too slow to be used in practice. The distilled student is significantly faster yet achieves competitive performance.}
\label{fig:teaser}
\end{figure}

To cope with these challenges, various egocentric action recognition methods \cite{radevski2021revisiting, herzig2022object, zhang2022object, materzynska2020something, wang2018videos, yan2020interactive, kim2021motion} demonstrate that explicitly modelling hand-object interactions (usually represented via bounding boxes \& object categories) significantly improves the action recognition performance, notably in a compositional generalization setup \cite{radevski2021revisiting}. Similarly, other works show that leveraging multiple modalities (optical flow, audio, etc.) at inference time yields improved performance \cite{xiong2022m, gabeur2020multi, kazakos2021little, nagrani2021attention}. The assumptions these methods make are (i) that all modalities used during training are also available at inference time, and (2) the compute budget at inference time would be sufficient to obtain and process the additional modalities. These assumptions make them cumbersome or even impossible to use in practice, e.g.\ on a limited compute budget, such as in the case of embedded devices. Namely, using dedicated models for each additional modality (e.g.\ object detector + tracker + transformer when using bounding boxes \& object categories \cite{radevski2021revisiting}), increases both the memory footprint as well as the inference time. 
\textit{Ideally, we leverage additional modalities only during training, while the resulting model uses RGB frames at inference time, i.e.\ when deployed in practice.}\par
One way to achieve the aforementioned goal is training Omnivorous models, i.e. models trained jointly on multiple modalities, which have been shown to generalize better \cite{girdhar2022omnivore}. In this work, we take a different route and transfer multimodal knowledge to models subsequently used in a unimodal setting. Namely, we distill the knowledge from a multimodal ensemble -- exhibiting superior performance, but unviable for deployment -- to a standard RGB-based action recognition model \cite{bertasius2021space} (See Fig.~\ref{fig:teaser}).\par
\textbf{Contributions.} We employ state-of-the-art knowledge distillation practices \cite{beyer2022knowledge} and \circled{1} show that a student~\cite{liu2021swin} taught by a multimodal teacher, is both more accurate and better calibrated than the same model trained from scratch or in an omnivorous fashion (\S\ref{sec:general-results}); \circled{2} We provide motivation and establish a simple but reliable multimodal distillation approach, which overcomes the issue of potentially suboptimal modality-specific teachers (\S\ref{sec:weighting}); \circled{3} We demonstrate that the distilled student performs on par with significantly larger models, and maintains performance in computationally cheaper inference setups (\S\ref{sec:speed}).\par
\section{Related Work}\label{sec:related_work}

\textbf{Knowledge distillation.} Originally introduced by Hinton \etal \cite{hinton2015distilling}, knowledge distillation is used to transfer the knowledge from one model, i.e.\ a teacher, to another model, i.e.\ a student, by training the student to match the teacher's (intermediate) outputs on a certain dataset. Shown to be useful in a variety of contexts, the primary goal is model compression - transferring the knowledge from a larger, cumbersome teacher model, or from a teacher exhibiting a different inductive bias \cite{touvron2021training, chen2022dearkd}, to a typically lightweight student model \cite{wang2021knowledge, cho2019efficacy, mishra2017apprentice}. Another line of work \cite{shen2020meal, asif2019ensemble} proposes to distill from ensembles of large teacher models to lightweight student models, obtaining promising results. Compared to these works, we focus on knowledge distillation from a multimodal teacher ensemble, i.e.\ a set of models each trained on a distinct modality.\par
\textbf{Multimodal knowledge distillation} has been previously used mainly in a cross-modal fashion, where the teacher and the student receive different modalities as input, e.g.\ the teacher receives RGB images while the student receives depth or optical flow images \cite{gupta2016cross}, or even a teacher which receives RGB images while the student receives audio as input \cite{aytar2016soundnet}. In contrast, other works \cite{xue2021multimodal} explore a multimodal knowledge expansion scenario, where a multimodal student learns from pseudo-labels of a unimodal teacher. We, on the other hand, focus on scenarios where obtaining additional modalities (optical flow, object detections, audio, etc.) during inference is prohibitive due to a limited compute budget, and therefore multimodal data is only leveraged during training time. Finally, multimodal knowledge distillation for action recognition has been previously explored by \cite{garcia2018modality, garcia2019dmcl}. They propose a hallucination stream, where given a modality A, a model is trained to imitate the (intermediate) features of a model trained on modality B. This approach is shown to yield performance improvements compared to an RGB baseline when using RGB and Depth data. In turn, the approach we adopt is the standard knowledge distillation \cite{hinton2015distilling}, following state-of-the-art distillation practices \cite{beyer2022knowledge}, featuring multiple different modalities such as RGB frames, optical flow, audio and object detections.\par

\textbf{Multimodal (egocentric) video understanding.} In the context of (egocentric) video understanding, several works have shown that using additional modalities at inference time significantly improves performance \cite{radevski2021revisiting, zhang2022object, herzig2022object, materzynska2020something, kim2021motion, nagrani2021attention, wang2018videos, tan2023egodistill}. The hypothesis is intuitive - certain actions are more easily understood from specific modalities, e.g.\ to recognize that a person is ``pushing something from left to right,'' the bounding boxes alone are sufficient \cite{radevski2021revisiting, materzynska2020something}. Nevertheless, the assumption these works make is that all modalities used during training are available during inference, or that the compute budget allows for processing additional modalities. Multiple works \cite{radevski2021revisiting, materzynska2020something, herzig2022object, wang2018videos, kim2021motion, zhang2022object} effectively use a Faster R-CNN \cite{ren2015faster}, tracker and object detection-specific models at inference time. In contrast, we posit that for egocentric video understanding, computing additional modalities on the fly may be prohibitive. Therefore, we propose a distillation approach which uses multiple modalities \textit{only} during training, while the resulting model uses only RGB frames for inference.\par

\textbf{Models robust to missing modalities during inference.} A parallel route to our goal is to explicitly train models to be robust to missing modalities during inference \cite{zhao2021missing, parthasarathy2020training, neverova2015moddrop}, or more recently, to process different modalities altogether interchangeably -- Omnivorous models \cite{girdhar2022omnivore, dai2022one, girdhar2022omnimae}. These models have been shown to generalize better than models trained on a single modality. In this work, we train an Omnivorous model using the same architecture as our student, and show that the student distilled from a multimodal teacher generalizes better than its Omnivorous variant.
\section{Methodology} \label{sec:methods}

\subsection{Egocentric Action Recognition}
% We assume we are given an input $\mathbf{x} \in \mathbb{R}^{T \times D_1 \times D_2, ..., \times D_L}$, which describes an egocentric action sequence, where $T$ is the number of time-steps, and for example, in the case of RGB frames $d_1 = H, d_2 = W, d_3 = C$, where $H$ is the frame height, $W$ is the frame width, and $C = 3$ are the channels.
We assume we are given an input $\mathbf{x} \in \mathbb{R}^{T \times D_1 \times D_2 ... \times D_L}$, which describes an egocentric action sequence, where $T$ is the number of time-steps, while $D_1 \times D_2 ... \times D_L$ represent other dimensions of the input data, e.g. the height, width and the number of channels of a video frame.
% \in \mathbb{R}^{T \times H \times W \times L}$
% which describes an egocentric action sequence
% , where $T$ is the number of sampled time-steps over the course of the action sequence,
% and $d_1, d_2, ..., d_L$ are the 
% $H$ and $W$ are the height and the width, and $L$ is the number of channels -- 3 for RGB, 2 for optical flow, and 1 for audio.
The goal of the model $f$ is to produce a discrete probability distribution over a predefined set of $C$ classes, i.e. $\mathbf{\hat{y}} = \sigma \left( f \left( \mathbf{x} \right) \right) \in \mathbb{R}_+^C$, where $\sigma$ is the softmax operator. The classes represent the actions, or alternatively, the nouns and verbs which constitute the actions (e.g. the object and the activity). 

Given a dataset $\mathbb{D} = \{ (\mathbf{x}_1, \mathbf{y}_1), \dots, (\mathbf{x}_{N}, \mathbf{y}_{N}) \}$ of $N$ egocentric action sequences $\mathbf{x}_i$ paired with labels $\mathbf{y}_i \in \mathbb{R}_+^C$, the model is trained by minimizing the standard cross-entropy objective $\mathcal{L}_{\text{CE}} = \frac{1}{N}\sum_{i=1}^{N} \mathbf{y}_i \cdot \log \sigma \left( f \left( \mathbf{x}_i \right) \right)$. In the case of compositional actions, characterized by separate nouns and verbs, and accompanied by their respective labels $\mathbf{y}^n_i$ and $\mathbf{y}^v_i$, separate prediction heads are used to produce $f^n(\mathbf{x}_i)$ and $f^v(\mathbf{x}_i)$. Finally, the model is trained by minimizing the sum of the loss terms corresponding to the nouns and verbs, $\mathcal{L}^n_{\text{CE}}$ and $\mathcal{L}^v_{\text{CE}}$ respectively.

% \subsection{Some Notes}
% Can we maybe formulate it like this?
% Different modalities - different manifestations of the unknown $\mathbf{x}$, i.-e. $\mathbf{x}^m = g^m(\mathbf{x})$, where $g^m$ is typically unknown.

% \begin{equation}
% \begin{gathered}
% \mathbf{\hat{y}^m_i} = f^m (\mathbf{x}^m_i) = f^m (g^m(\mathbf{x}_i)),
% \end{gathered}
% \end{equation}

% \begin{equation}
%     p(\mathbf{x}) = p ( \mathbf{x}^0, \mathbf{x}^1, \dots,  \mathbf{x}^{T - 1} ) = \sum p (\mathbf{x}^m) = \sum p( g^m (\mathbf{x}) )
% \end{equation}

% \begin{equation}
%     p(\mathbf{x}) = \int p( \mathbf{x} | g ) dg 
% \end{equation}

% \begin{equation}
% \begin{gathered}
% P(\mathbf{y} | \mathbf{x}) = f^m (\mathbf{x}^m_i) = f^m (g^m(\mathbf{x}_i)),
% \end{gathered}
% \end{equation}

\subsection{Multimodal Knowledge Distillation}\label{sec:mmkd}

In egocentric vision, there often exist multiple input modalities that characterize the same actions. The action recognition task may thus be performed via the use of multiple modalities, leveraged both during training \cite{girdhar2022omnivore} and inference \cite{xiong2022m, gabeur2020multi, kazakos2021little, nagrani2021attention} by e.g. ensembling \cite{xiong2022m}, multimodal-fusion \cite{radevski2021revisiting, kazakos2021little}, etc. However, in the case of the latter, processing multimodal data may be computationally prohibitive at inference time.\par 
The fundamental concept our method builds on is knowledge distillation \cite{hinton2015distilling}, featuring a teacher (usually larger model, exhibiting strong performance, but cumbersome to use in practice) and a student (typically a smaller model, trained to mimic the teacher \cite{beyer2022knowledge}).
Focusing on the most accessible data modality -- RGB video frames (e.g. obtained using a single monocular video camera) -- we opt for distilling the knowledge of a multimodal ensemble to a single model that relies on RGB inputs alone. We make a modification to the standard knowledge distillation approach, by altering the teacher such that (i) it is not a single model, but rather an ensemble of models, and (ii) the constituting models get different modalities as input.\par
\textbf{Teacher ensemble.} Given $M$ datasets $\mathbb{D}^m = \{ (\mathbf{x}^m_1, \mathbf{y}_1), \dots, (\mathbf{x}^m_{N}, \mathbf{y}_{N}) \}$ of different modalities, we train a separate model $f^m$ by optimizing the learning objective $\mathcal{L}^m_{\text{CE}} = \frac{1}{N} \sum_{i=1}^{N} \mathbf{y}_i \cdot \log \sigma \left( f^m \left( \mathbf{x}^m_i \right) \right)$ for each modality. Finally, the ensemble output can be obtained by averaging the teacher outputs, i.e. $\mathbf{\hat{y}}^t_i = \sigma \left( \frac{1}{M}\sum_{m=1}^{M} f^m (\mathbf{x}_i^m) \right)$.\par
% of the individual teachers:
% \begin{equation}
% \begin{gathered}
% \mathbf{\hat{y}}^t_i = \sigma \left( \frac{1}{M}\sum_{m=1}^{M} f^m (\mathbf{x}_i^m) \right).
% \end{gathered}
% \label{eq:ensemble-output}
% \end{equation}
 Intuitively, under-performing modality-specific models could negatively affect the performance of the ensemble. We thus consider assigning different weights to the output logits of each model in the ensemble, before aggregating their predictions. Ideally, we would want to perform a Bayesian prediction, i.e.\ $p( \mathbf{y} | \mathbf{x}, \mathbb{D}) = \int_{f \in F} p( \mathbf{y} | \mathbf{x} , f) p( f | \mathbb{D}) df$.
% \begin{align}
%     p( \mathbf{y} | \mathbf{x}, \mathbb{D}) &= \int_{f \in F} p( \mathbf{y} | \mathbf{x} , f) p( f | \mathbb{D}) df
% \label{eq:bayesian}
% \end{align}
For a given finite ensemble of $M$ diverse predictors, we replace the integral with a sum over the individual models, i.e.\ $p( \mathbf{y} | \mathbf{x}, \mathbb{D}) \approx \sum_{m=1}^{M} p( \mathbf{y} | \mathbf{x} , f^m) p( f^m | \mathbb{D})$.
% \begin{align}
%     p( \mathbf{y} | \mathbf{x}, \mathbb{D}) \approx \sum_{m=1}^{M} p( \mathbf{y} | \mathbf{x} , f^m) p( f^m | \mathbb{D})
% \label{eq:bayesian}
% \end{align}
We further approximate $p( f^m | \mathbb{D})$ via its proportionality to the data likelihood $p( \mathbb{D} | f^m)$ under Bayes rule, i.e. $p( f^m | \mathbb{D}) \propto p( \mathbb{D} | f^m)$, which itself can be expressed in terms of the cross-entropy that the model $f^m$ exhibits on the dataset $\mathbb{D}$. 

The cross-entropy $e^m$ of each modality-specific model in the ensemble can be estimated as the average over a holdout set, i.e. $e^m = \frac{1}{Z}\sum_{i=1}^{Z} \text{CE}(f^m(\mathbf{x}^m_i), \mathbf{y}_i)$, where $Z$ is the number of held-out samples used to estimate the weights. Then, we can obtain the weights for the modality-specific models via softmax normalization of negative cross-entropy terms, i.e. $w^m \propto \text{exp}(-e^m/\gamma)$, where $\gamma$ is a temperature term which controls entropy of the model weights, e.g. if $\gamma \rightarrow \infty$, equal weights would be given to each teacher -- resulting in an arithmetic average. 
% Given a data sample with ground truth label $\mathbf{y}_i$, we obtain a per-teacher error $e^m$ as the cross-entropy w.r.t.\ the ground truth labels $e^m = \text{CE}(f(\mathbf{x}^m_i), \mathbf{y}_i)$, where $m$ indicates the modality. 
% Then, we obtain the weights for each of the modality-specific models in the ensemble as $w^{m} = \sfrac{\exp(-\sfrac{e^m}{\gamma})}{\sum_{j=1}^{M} \exp(-\sfrac{e^j}{\gamma})}$, 
% % \begin{equation}
% % \begin{gathered}
% %     w^{m} = \sfrac{\exp(-\sfrac{e^m}{\gamma})}{\sum_{j=1}^{M} \exp(-\sfrac{e^j}{\gamma})},
% % \end{gathered}
% % \end{equation}
% where $M$ is the number of modalities \& modality-specific teachers, and $\gamma$ is a temperature term which controls entropy of the weights, i.e., if $\gamma \rightarrow \infty$, equal weights would be given to each teacher -- resulting in an unweighted average. The per-teacher error $e^m$ can be estimated on a per-sample basis during training, or obtained as the average error over a holdout set: $e^m = \frac{1}{Z}\sum_{i=1}^{Z} \text{CE}(f(\mathbf{x}^m_i), \mathbf{y}_i)$, where $Z$ is the number of held-out samples used to estimate the weights. 
We finally compute the weighted average of the predictions of $M$ modality-specific models as the teacher output:

\begin{equation}
\begin{gathered}
\mathbf{\hat{y}}^t_i = \sigma \left( \sum_{m=1}^{M} w^m f^m (\mathbf{x}_i^m) \right).
\end{gathered}
\end{equation}

\begin{figure}[t]
\centering
\includegraphics[width=1.0\columnwidth]{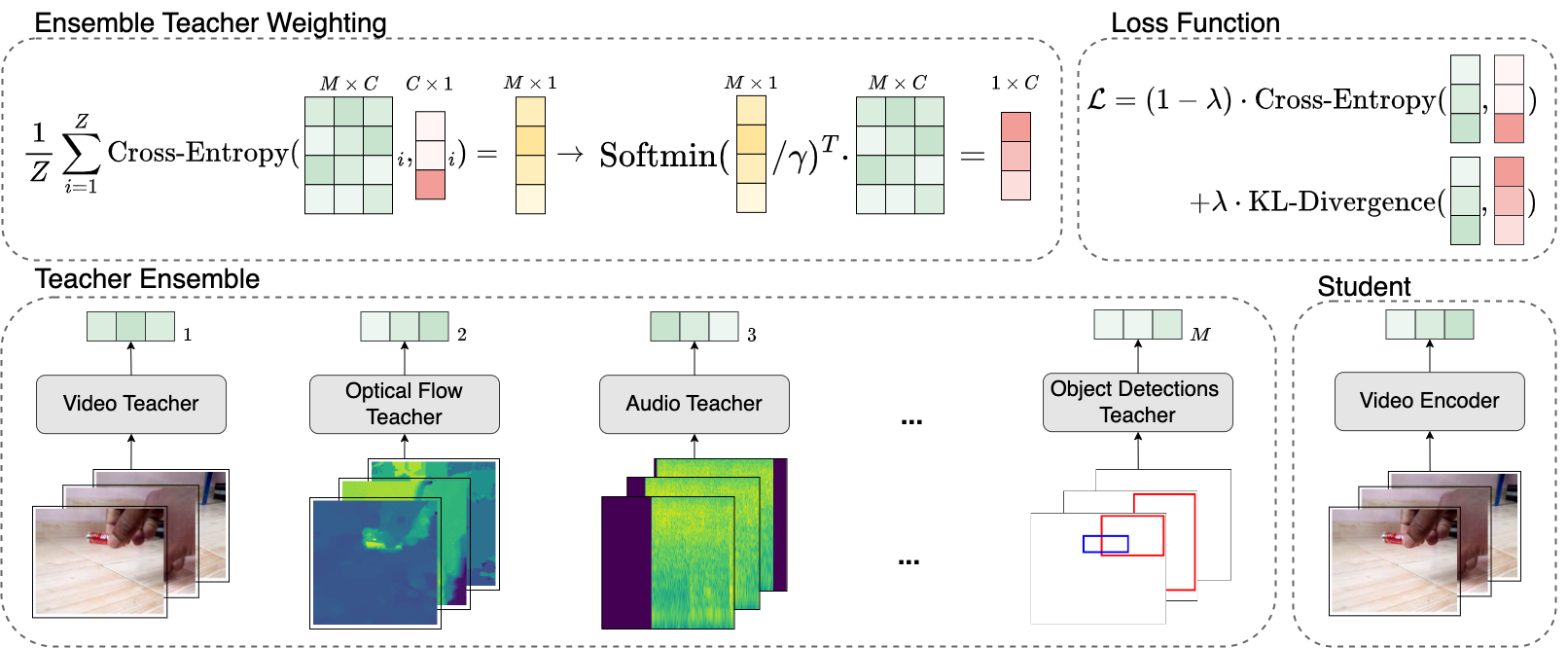}
\caption{Overview of the main components of our approach. }
\label{fig:mainfigure}
\end{figure}

Figure \ref{fig:mainfigure} presents a high-level overview of our approach. In summary, our student is taught by a multimodal teacher which is itself an ensemble of multiple modality-specific models, trained separately on each modality.\par
\textbf{Training objective.} During training we perform multimodal knowledge distillation, as originally proposed by Hinton \etal \cite{hinton2015distilling}. To be specific, we minimize the KL-divergence $\mathcal{L}_{\text{KL}}$ between the class probabilities predicted by the teacher $\mathbf{\hat{y}}^t_i = [\hat{y}^t_{i,1}, \dots, \hat{y}^t_{i,C}] \in \mathbb{R}_+^C$, and the student $\mathbf{\hat{y}}^s_i = [\hat{y}^s_{i,1}, \dots, \hat{y}^s_{i, C}] \in \mathbb{R}_+^C$ as $\mathcal{L}_{\text{KL}} = \frac{1}{N}\sum_{i=1}^{N} (- \mathbf{\hat{y}}^t_i \cdot \log \mathbf{\hat{y}}^s_i + \mathbf{\hat{y}}^t_i\cdot \log \mathbf{\hat{y}}^t_i)$.

% \begin{equation}
% \begin{gathered}
%     \mathcal{L}_{\text{KL}} = \frac{1}{N}\sum_{i=1}^{N} (- \mathbf{\hat{y}}^t_i \cdot \log \mathbf{\hat{y}}^s_i + \mathbf{\hat{y}}^t_i\cdot \log \mathbf{\hat{y}}^t_i)
% \end{gathered}
% \label{eq:kl_divergence}
% \end{equation}

Additionally, we use a temperature parameter $\tau$ to control the entropy of the predicted probability scores while preserving their ranking $\hat{y}^s_j \propto \text{exp}(\hat{y}^s_j/\tau)$. As per standard practice \cite{hinton2015distilling}, we rescale the KL-divergence loss based on the temperature by $\mathcal{L}_{\text{KL}} = \mathcal{L}_{\text{KL}} \cdot \tau^2$. We further use the cross-entropy action recognition loss $\mathcal{L}_{\text{CE}}$, and compute the final loss $\mathcal{L} = \lambda \cdot \mathcal{L}_{\text{KL}} + (1 - \lambda) \cdot \mathcal{L}_{\text{CE}}$, 
% \begin{equation}
% \begin{gathered}
%  \mathcal{L} = \lambda \cdot \mathcal{L}_{\text{KL}} + (1 - \lambda) \cdot \mathcal{L}_{\text{CE}},
% \end{gathered}
% \end{equation}
where $\lambda$ balances the distillation loss $\mathcal{L}_{\text{KL}}$ and the action recognition loss $\mathcal{L}_{\text{CE}}$. For example, with $\lambda = 0.0$ we would effectively be training the modality-specific RGB model, while with $\lambda = 1.0$, we would perform solely multimodal knowledge distillation.

% Show that the teacher logits weighted by the risk of individual teachers would yield outputs closer to the true Bayesian class probabilities than individual teacher or a simple average. 

% \begin{align}
%     p( \mathbf{y} | \mathbf{x}, \mathbb{D}) = \int_{f \in F} p( \mathbf{y} | \mathbf{x} , f) p( f | \mathbb{D}) df
% \end{align}

% Using a fixed number of models $f_i \in \{ f_1, ..., f_T \}$, we obtain the prediction of an ensemble as:
% \begin{align}
%     f( \mathbf{x}) = \sum_{i=1}^{T} w_i f_i ( \mathbf{x} ) = \sum_{i=1}^{T} w_i p( \mathbf{y} | \mathbf{x} , f_i)
% \end{align}

% where the model weights $w_i$ are estimated as the average risk of each of the individual models measured on the held-out dataset as $w_i = \mathbb{E}_{\mathbf{x},\mathbf{y} \sim  \mathbb{D}} \left [ l \left(\mathbf{y}, f_i(\mathbf{x}) \right) \right ]$ 
% % $\propto p(f_i | \mathbb{D}) = \mathbb{E}_{\mathbf{x},\mathbf{y} \sim  \mathbb{D}} p(f_i | \mathbf{x}, \mathbf{y})$, 
% such that they reflect the $p( f_i | \mathbb{D})$ of individual models.

\textbf{Optimality of a multimodal ensemble teacher.} The work of Menon \etal \cite{menon2021statistical} further demonstrates that in the context of distillation, a teacher that predicts the Bayes class-probability distribution over the labels $\mathbf{p}^*(\mathbf{x}) = [\mathbb{P}(y|\mathbf{x})]_{y \in [C]}$ exhibits the lowest possible variance of the student objective for any convex loss function.
% $l$:

% \begin{align}
%     & \nonumber \mathbb{V}_{{\{\mathbf{x_n},\mathbf{y_n}\}}_{n=1}^N \sim P^N} \left [ \mathbb{E}_{\mathbf{x}} \left [ l \left ( \mathbf{p}^*(\mathbf{x}), f(\mathbf{x}) \right) \right ] \right] \\ & \leq \mathbb{V}_{{\{\mathbf{x_n},\mathbf{y_n}\}}_{n=1}^N \sim P^N} \left [ \mathbb{E}_{\mathbf{x}, \mathbf{y}} \left [  l \left(\mathbf{y},  f(\mathbf{x}) \right) \right ] \right ]
% \end{align}

The student trained to minimize the KL-divergence between its output and the Bayes class-probabilities would thus generalize the best. In a preliminary experiment shown in Table~\ref{table:calibration-error-teacher}, we demonstrate that the multimodal ensemble (both for $\gamma=1$ and $\gamma=30.0$) achieves a significantly higher accuracy and lower calibration error, and thus represents a better approximation of the Bayes class probabilities than a single modality-specific model. This may lead to
a lower-variance objective for the student and improved generalization during knowledge distillation \cite{menon2021statistical}.
% (see \S\ref{sec:experiments} for details on the datasets, models and metrics).
% We observe that the proposed weighted averaging of model outputs ($\gamma \to\infty$) in the ensemble achieves both a higher accuracy, as well as the lower calibration error than other models.

\begin{table}[t]
\centering
\resizebox{1.0\columnwidth}{!}{
\begin{tabular}{>{\columncolor{graygray}}c>{\columncolor{greengreen}}c>{\columncolor{greengreen}}c>{\columncolor{greengreen}}c>{\columncolor{greengreen}}c>{\columncolor{blueblue}}c>{\columncolor{blueblue}}c} \toprule
    \cellcolor{white}{} & \multicolumn{4}{c}{Epic-Kitchens Regular} & \multicolumn{2}{c}{Something-Something} \\ \midrule
    {Method} & {Noun Acc.} & {Noun ECE} & {Verb Acc.} & {Verb ECE} & {Action Acc.} & {Action ECE} \\ \midrule
    RGB Baseline & 52.0 & 19.9 & 61.7 & 17.1 & 59.3 & 21.2 \\
    Teacher ($\gamma = 30.0$) & 52.3 & 2.04 & 66.8 & 3.4 & 66.6 & 8.1 \\
    Teacher ($\gamma = 1.0$) & 53.9 & 8.9 & 66.9 & 4.2 & 66.7 & 9.0 \\ \midrule
    \cellcolor{white}{} & \multicolumn{4}{c}{Epic-Kitchens Unseen} & \multicolumn{2}{c}{Something-Else} \\ \midrule
    {Method} & {Noun Acc.} & {Noun ECE} & {Verb Acc.} & {Verb ECE} & {Action Acc.} & {Action ECE} \\  \midrule
    RGB Baseline & 38.3 & 26.9 & 51.7 & 24.0 & 51.8 & 21.4 \\
    Teacher ($\gamma = 30.0$) & 42.9 & 6.02 & 58.0 & 9.9 & 63.3 & 6.5 \\
    Teacher ($\gamma = 1.0$) & 43.8 & 9.5 & 57.7 & 9.7 & 63.5 & 8.1 \\ \bottomrule
\end{tabular}
}
\caption{Accuracy \& Expected Calibration Error (ECE) of the RGB baseline and Teacher ensemble on Epic-Kitchens and Something-Something (regular \& unseen splits). All available modalities for the respective dataset used in the ensemble.}
\label{table:calibration-error-teacher}
\end{table}

\section{Experiments \& Discussion}\label{sec:experiments}

\textbf{Datasets.} We use Something-Something (V2) \cite{goyal2017something} and Epic-Kitchens (100) \cite{damen2020rescaling}. Something-Something contains videos of people performing 174 (object agnostic) unique actions with their hands, e.g. ``pushing [something] left'', ``taking [something] out of [something]'', etc. Epic-Kitchens' videos take place in kitchen environments, where the actions are noun-verb compositions. The 300 unique nouns indicate the active object in the video, while 97 unique verbs indicate the activity, e.g. ``cutting carrot'', ``washing pan'', etc. The action is considered correct if both the noun and the verb are correctly predicted. Additionally, we use the Something-Else \cite{materzynska2020something} and the Epic-Kitchens Unseen split to measure the compositional generalization ability of the models w.r.t.\ unseen environments \& objects.\par

\textbf{Modalities.} In addition to the RGB frames (the modality of interest at inference time), we consider the following:
\begin{inparaenum}[(i)]
    \item \textit{Optical flow} (OF). First used by I3D \cite{carreira2017quo} for action recognition. We use optical flow obtained using TV-L\textsubscript{1}\cite{zach2007duality}.
    \item \textit{Object detections} (OBJ). Shown to significantly improve the performance of standard (RGB) egocentric video understanding models across datasets \cite{radevski2021revisiting, herzig2022object, materzynska2020something, zhang2022object, wang2018videos}. 
    % The object detector is usually trained on object-agnostic annotations \cite{shan2020understanding}, i.e., it can recognize two ``hands'' and ``objects''. 
    As per \cite{shan2020understanding}, we use the object detector trained on object-agnostic annotations, i.e. with ``hands'' and ``objects'' object labels.
    \item \textit{Audio} (A). For certain datasets \cite{damen2018scaling, damen2020rescaling}, several works \cite{kazakos2021little, xiong2022m} have shown that using audio improves action recognition performance. The audio is obtained directly by the participant recording the action.
\end{inparaenum}\par
Moreover, if available, one may include additional modalities, e.g.\ depth estimates. For Something-Something/Else the audio is unavailable, and we use the object detections provided by \cite{herzig2022object, materzynska2020something}, and optical flow from \cite{wang2016temporal}. For Epic-Kitchens we use the modalities available and released with the dataset \cite{damen2020rescaling}.\par

\textbf{Models.} We use a Swin-T Transformer \cite{liu2021swin} to encode RGB frames, optical flow, and audio. Each optical flow frame is represented as a $224 \times224 \times 2$ tensor, where the two values at each spatial location represent the $x$ and $y$ velocity components. In case of audio, we extract 1.116 second-audio segments (0.558s before its corresponding time-step and 0.558s after it) for each frame. We compute the mel-spectrogram of the audio segment (details in Supp.), which is subsequently resized to desired width and height.
%, using 1024 FFT bins, 128 mel filterbanks, Hann window and the window length of 160 and the hop length of 80 audio samples.
We thus treat each modality as a sequence of $224 \times 224$ multi-channel images, which we provide as input to the vision transformer, a common procedure in recent vision models \cite{girdhar2022omnimae, girdhar2022omnivore}. To encode the object detections, i.e.\ bounding boxes and object categories of the scene objects, we use a state-of-the-art model - STLT \cite{radevski2021revisiting}. In STLT, a spatial and temporal transformer separately encode the spatial and temporal arrangement of the objects occurring in the video. Our multimodal teacher is an ensemble of individual, modality-specific models.
% , which aggregates the logits of each model in the ensemble.
Unless noted otherwise, the student is a Swin-T model which receives RGB frames as input, both during training (distillation) and inference.
In addition to Swin-T, in \S\ref{sec:speed}, we also consider ResNet3D \cite{kataoka2020would} (18 and 50 layers deep) models which receive video frames of size $112 \times 112$ as input as light-weight students.

\textbf{Metrics.}
Besides accuracy, we measure Expected Calibration Error (ECE) \cite{guo2017calibration}. As per \cite{guo2017calibration, rousseau2021post}, we sort the predictions based on the per-class confidence scores and group them into $K$ bins $B_k$, each associated with a confidence interval $I_{B_k} = (\frac{k-1}{K} ,\frac{k}{K})$, where $K=15$. ECE represents the discrepancy between the average accuracy $acc(B_k)$ and the average confidence $conf(B_k)$ in each bin $B_k$, i.e. $\text{ECE} = \sum_{k=1}^{K} \frac{|B_k|}{N} |acc(B_k) - conf(B_k)|$, where $N$ is the number of evaluation samples.\par
% \begin{equation}
% \begin{gathered}
%     \text{ECE} = \sum_{k=1}^{K} \frac{|B_k|}{N} |acc(B_k) - conf(B_k)|,
% \end{gathered}
% \end{equation}

% \textbf{Baselines.} In the experiments that follow, we compare our proposed method against:
% \begin{itemize}
%     \item RGB baseline \cite{liu2021swin}, trained individually only the RGB frames;
%     \item Modality-specific models, trained on a single modality other than the RGB modality;
%     \item Teachers, ensembles by logits average of modality-specific models including the RGB model;
%     \item Omnivore \cite{girdhar2022omnivore}, trained  \textit{M}x longer that the other models, on \textit{all M modalities} jointly by randomly sampling a modality for each video during training.
% \end{itemize}
\textbf{Implementation details.} We train all models for 60 epochs using AdamW \cite{loshchilov2017decoupled}, with a peak learning rate of $1e-4$, linearly increased for the first 5\% of the training and decreased to 0.0 till the end, weight decay of $5e-2$, and clip the gradients when the norm exceeds 5.0. For Epic-Kitchens, we sample 32 frames with a fixed stride of 2, and for Something-Something we evenly sample 16 frames to cover the whole video. We use a single spatial and temporal crop, unless stated otherwise. During training, we chose a random start frame, while during inference, we select the start frame such that the sequence covers the central portion of the video. In the case of multiple temporal crops as test-time augmentation, we chose the start frames such that the video is covered uniformly. During training we apply standard data augmentations -- random spatial video crops, color jittering, and horizontal flips (for Epic-Kitchens only). The temperature parameter $\tau$ is fixed to 10.0 for both the student and the teacher. In \S\ref{sec:weighting} we ablate the impact of the loss balancing term $\lambda$ and the Ensemble Teacher Weighting temperature term $\gamma$\footnote{As the datasets test sets either do not exist \cite{materzynska2020something}, or have restricted access, we report results using the model after the final training epoch.}.\par
During training we follow the consistent teaching paradigm \cite{beyer2022knowledge} where the student and teacher strictly receive the same views of the data -- we ensure for spatial and temporal consistency, i.e.\  the models receive the same frame indices, same random crops, and horizontal flips.

\subsection{Multimodal Distillation for Egocentric Vision}\label{sec:general-results}
We first verify the overall effectiveness of multimodal knowledge distillation on the task of egocentric action recognition for both object agnostic actions (Something-Something) and actions as noun-verb compositions (Epic-Kitchens). Across all experiments, we fix $\lambda$ to 1.0, i.e. we train solely with multimodal knowledge distillation. Similarly, we set $\gamma$ to a large value ($\gamma = 30.0$), where effectively each teacher equally contributes to the ensemble output.
% , i.e., $w^m = 1/M$, where $M$ is the number of modality-specific teachers.
Note that in this setting, modalities such as optical flow and audio may underperform and adversely affect the performance of recognizing active objects, i.e.\ nouns.\par

\subsubsection{Recognizing Egocentric Actions}

\begin{table*}[t]
\begin{subtable}[t]{0.48\textwidth}
\centering
\resizebox{\columnwidth}{!}{
\begin{tabular}{>{\columncolor{graygray}}c>{\columncolor{pinkpink}}c>{\columncolor{yellowyellow}}c>{\columncolor{greengreen}}c>{\columncolor{greengreen}}c>{\columncolor{greengreen}}c} \toprule
    {Method} & {Training Modalities} & {Inference Modalities} & {Noun} & {Verb} & {Action} \\ \midrule
    Baseline & RGB & RGB & 52.0 & 61.7 & 38.3 \\ \midrule \midrule
    Modality-specific & OF & OF & 34.1 & 59.0 & 25.9 \\
    Teacher & \NA & RGB \& OF & 51.9 & 65.3 & 39.5 \\
    Student & RGB \& OF & RGB & 52.2\textsubscript{\textcolor{redred}{\textbf{+0.2}}} & 65.6\textsubscript{\textcolor{redred}{\textbf{+3.9}}} & 39.9\textsubscript{\textcolor{redred}{\textbf{+1.6}}} \\ \midrule
    Modality-specific & A & A & 22.3 & 46.5 & 15.1 \\
    Teacher & \NA & RGB \& A & 52.7 & 64.4 & 39.8 \\
    Student & RGB \& A & RGB & 51.5\textsubscript{\textcolor{redred}{\textbf{-0.5}}} & 62.4\textsubscript{\textcolor{redred}{\textbf{+0.7}}} & 37.9\textsubscript{\textcolor{redred}{\textbf{-0.4}}} \\ \midrule
  Teacher & \NA & RGB \& OF \& A & 52.3 & 66.8 & 40.5 \\
    Student & RGB \& OF \& A & RGB & 51.7\textsubscript{\textcolor{redred}{\textbf{-0.3}}} & 65.4\textsubscript{\textcolor{redred}{\textbf{+3.7}}} & 39.3\textsubscript{\textcolor{redred}{\textbf{+1.0}}} \\ \bottomrule
\end{tabular}
}
\caption{Epic-Kitchens \cite{damen2020rescaling} active object (noun) and activity (verb) recognition.}
\label{table:epic-regular}
\end{subtable}
\hfill
\begin{subtable}[t]{0.48\textwidth}
\centering
\resizebox{0.9\columnwidth}{!}{
\begin{tabular}{>{\columncolor{graygray}}c>{\columncolor{pinkpink}}c>{\columncolor{yellowyellow}}c>{\columncolor{blueblue}}c>{\columncolor{blueblue}}c} \toprule
    {Method} & {Training Modalities} & {Inference Modalities} & {Action@1} & {Action@5} \\ \midrule
    Baseline & RGB & RGB & 60.3 & 86.4 \\ \midrule \midrule
    Modality-specific & OF & OF & 49.3 & 79.0 \\
    Teacher & \NA & RGB \& OF & 64.3 & 88.9 \\
    Student & RGB \& OF & RGB & 62.8\textsubscript{\textcolor{redred}{\textbf{+2.5}}} & 88.9\textsubscript{\textcolor{redred}{\textbf{+2.5}}} \\ \midrule
    Modality-specific & OBJ & OBJ & 47.9 & 76.2 \\
    Teacher & \NA & RGB \& OBJ & 65.3 & 89.5 \\
    Student & RGB \& OBJ & RGB & 63.2\textsubscript{\textcolor{redred}{\textbf{+2.9}}} & 88.7\textsubscript{\textcolor{redred}{\textbf{+2.3}}} \\ \midrule
    Teacher & \NA & RGB \& OF \& OBJ & 66.6 & 90.5 \\
    Student & RGB \& OF \& OBJ & RGB & 63.0\textsubscript{\textcolor{redred}{\textbf{+2.7}}} & 88.9\textsubscript{\textcolor{redred}{\textbf{+2.5}}} \\
    \bottomrule
\end{tabular}
}
\caption{Something-Something \cite{goyal2017something} object-agnostic action recognition.}
\label{table:something-regular}
\end{subtable}
\caption{Egocentric action recognition. RGB = Video frames; OF = Optical flow; A = Audio; OBJ = Object detections. Multimodal distillation with $\lambda = 1.0$ and $\gamma = 30.0$. Improvement over RGB frames baseline \cite{liu2021swin} in \textbf{\textcolor{redred}{red}}.}
\label{table:regular-actions}
\end{table*}

In Table~\ref{table:regular-actions}, we report performance on Something-Something V2 \cite{goyal2017something} and Epic-Kitchens 100 \cite{damen2020rescaling}. As reported by others \cite{radevski2021revisiting, materzynska2020something, xiong2022m}, employing additional modalities at inference time significantly improves the performance compared to the RGB baseline model, for both Epic-Kitchens and  Something-Something.\par
% Nevertheless, we make the assumption that computing optical flow, or running object detectors at inference time is not feasible due to a limited compute budget.
A novel observation by our work is that \textit{multimodal knowledge distillation performs well in the context of egocentric video understanding}. For Epic-Kitchens (Table~\ref{table:epic-regular}) we observe that when the student is distilled from an RGB \& OF, or RGB \& OF \& A teacher, it is superior to the baseline model, as well as all modality-specific models, for recognizing actions. On the other hand, distilling from an RGB \& A teacher yields performance lower than the baseline, due to the low performance of the Audio model itself. Specifically, we observe that the audio-specific teacher lowers the noun (active object) recognition performance. In \S\ref{sec:weighting} we propose a solution for this issue.
For Something-Something (Table~\ref{table:something-regular}), the student model is superior to the baseline for each combination of modalities. When distilling from all available modalities (RGB \& OF \& OBJ), the resulting model outperforms the baseline by \textcolor{redred}{\textbf{3.7\%}} in terms of top-1 accuracy. In terms of top-5 accuracy on Something-Something, the students achieve performance that is nearly on par with the teacher.

\subsubsection{Generalizing to Unseen Environments \& Objects}
We investigate to what extent our findings translate to the compositional generalization setting, in which the performance of standard video models deteriorates significantly \cite{materzynska2020something}. We use the Epic-Kitchens Unseen split and the Something-Else \cite{materzynska2020something} split. On Something-Something the objects the people interact with might overlap between training and testing, potentially allowing models to pick up undesirable biases w.r.t.\  object appearance to discriminate between the actions. Therefore, Something-Else proposes a compositional generalization data split, on which the objects at training and testing time do not overlap, such that the models encounter strictly novel objects during testing. Likewise, the Epic-Kitchens Unseen split selects videos from participants unobserved in the training data.\par

\begin{table*}[t]
\begin{subtable}[t]{0.48\textwidth}
\centering
\resizebox{\columnwidth}{!}{
\begin{tabular}{>{\columncolor{graygray}}c>{\columncolor{pinkpink}}c>{\columncolor{yellowyellow}}c>{\columncolor{greengreen}}c>{\columncolor{greengreen}}c>{\columncolor{greengreen}}c} \toprule
    {Method} & {Training Modalities} & {Inference Modalities} & {Noun} & {Verb} & {Action} \\ \midrule
    Baseline & RGB & RGB & 38.3 & 51.7 & 25.4 \\ \midrule \midrule
    Modality-specific & OF & OF & 28.0 & 53.2 & 21.6 \\
    Teacher & \NA & RGB \& OF & 41.0 & 54.9 & 28.4 \\
    Student & RGB \& OF & RGB & 42.5\textsubscript{\textcolor{redred}{\textbf{+4.2}}} & 55.9\textsubscript{\textcolor{redred}{\textbf{+4.2}}} & 30.2\textsubscript{\textcolor{redred}{\textbf{+4.8}}} \\ \midrule
    Modality-specific & A & A & 15.0 & 41.5 & 9.1 \\
    Teacher & \NA & RGB \& A & 41.9 & 55.3 & 28.5 \\
    Student & RGB \& A & RGB & 41.8\textsubscript{\textcolor{redred}{\textbf{+3.5}}} & 51.8\textsubscript{\textcolor{redred}{\textbf{+0.1}}} & 27.5\textsubscript{\textcolor{redred}{\textbf{+2.1}}} \\ \midrule
  Teacher & \NA & RGB \& OF \& A & 42.9 & 58.0 & 30.3 \\
    Student & RGB \& OF \& A & RGB & 43.7\textsubscript{\textcolor{redred}{\textbf{+5.4}}} & 54.1\textsubscript{\textcolor{redred}{\textbf{+3.4}}} & 29.6\textsubscript{\textcolor{redred}{\textbf{+4.2}}} \\ \bottomrule
\end{tabular}
}
\caption{Epic-Kitchens \cite{damen2020rescaling} active object (noun) and activity (verb) recognition on participants unseen during training.}
\label{table:unseen-epic}
\end{subtable}
\hfill
\begin{subtable}[t]{0.48\textwidth}
\centering
\resizebox{0.9\columnwidth}{!}{
\begin{tabular}{>{\columncolor{graygray}}c>{\columncolor{pinkpink}}c>{\columncolor{yellowyellow}}c>{\columncolor{blueblue}}c>{\columncolor{blueblue}}c} \toprule
    {Method} & {Training Modalities} & {Inference Modalities} & {Action@1} & {Action@5} \\ \midrule
    Baseline & RGB & RGB & 51.8 & 79.5 \\ \midrule \midrule
    Modality-specific & OF & OF & 49.0 & 77.4 \\
    Teacher & \NA & RGB \& OF & 61.0 & 86.4 \\
    Student & RGB \& OF & RGB & 58.2\textsubscript{\textcolor{redred}{\textbf{+6.4}}} & 85.1\textsubscript{\textcolor{redred}{\textbf{+5.6}}} \\ \midrule
    Modality-specific & OBJ & OBJ & 41.4 & 67.3 \\
    Teacher & \NA & RGB \& OBJ & 59.4 & 84.5 \\
    Student & RGB \& OBJ & RGB & 57.5\textsubscript{\textcolor{redred}{\textbf{+5.7}}} & 84.1\textsubscript{\textcolor{redred}{\textbf{+4.6}}} \\ \midrule
    Teacher & \NA & RGB \& OF \& OBJ & 63.6 & 87.7 \\
    Student & RGB \& OF \& OBJ & RGB & 59.1\textsubscript{\textcolor{redred}{\textbf{+7.3}}} & 86.1\textsubscript{\textcolor{redred}{\textbf{+6.6}}} \\
    \bottomrule
\end{tabular}
}
\caption{Something-Else \cite{materzynska2020something} object-agnostic action recognition featuring objects unseen during training.}
\label{table:unseen-something}
\end{subtable}
\caption{Egocentric action recognition with unseen environments and objects. RGB = Video frames; OF = Optical flow; A = Audio; OBJ = Object detections. Multimodal distillation with $\lambda = 1.0$ and $\gamma = 30.0$. Improvement over RGB frames baseline \cite{liu2021swin} in \textbf{\textcolor{redred}{red}.}}
\label{table:unseen-actions}
\end{table*}

We report performance in Table~\ref{table:unseen-actions}. The general observation across datasets and modalities is that the students significantly outperform the RGB baseline model, and are sometimes even competitive with the teacher. Notably, on Something-Else, the student distilled from an RGB \& OF \& OBJ teacher outperforms the baseline by \textcolor{redred}{\textbf{7.3\%}} in terms of top-1 accuracy.

\subsubsection{Comparison with Omnivorous Models}
We explicitly compare multimodal knowledge distillation against Omnivorous models \cite{girdhar2022omnivore, girdhar2022omnimae}, which are trained jointly on all modalities, originally only on non-aligned data, e.g.\ training on RGB frames from one dataset and depth maps from another \cite{girdhar2022omnivore}. These models have been proven to generalize better compared to their unimodal counterparts. We train the Omnivore model using batches comprised of data from the same modality, where we randomly sample the modality for each batch \footnote{Girdhar \etal \cite{girdhar2022omnivore} show that there is no performance difference if the batches contain mixed data from different modalities, however, we found that using homogeneous batches yields better performance.}. Additionally, we train the Omnivore models $M \times$ more epochs ($M$ is the number of modalities), to account for the random sampling of modalities during training. We represent the object detections modality as bounding boxes (with the line thickness of 2px) pasted on a white canvas where each category is colored uniquely, e.g.\ hand in blue and object in red.\par
We report results in Table~\ref{table:omnivorous-models}, and observe that while on the Epic-Kitchens Unseen split and the Something-Else split the Omnivore model exhibits strong performance ($\sim$5.5\% improvement on top of the baseline on Something-Else), the multimodal distillation approach achieves even higher performance. However, we conclude that both approaches represent viable options for training multimodal models which leverage unimodal inputs during inference.

\begin{table}[t]
\centering
\resizebox{0.9\columnwidth}{!}{
\begin{tabular}{>{\columncolor{graygray}}c>{\columncolor{greengreen}}c>{\columncolor{greengreen}}c>{\columncolor{greengreen}}c>{\columncolor{blueblue}}c>{\columncolor{blueblue}}c} \toprule
    \cellcolor{white}{} & \multicolumn{3}{c}{Epic-Kitchens Regular split} & \multicolumn{2}{c}{Something-Something} \\ \midrule
    {Method} & {Noun@1} & {Verb@1} & {Action@1} & {Action@1} & {Action@5} \\  \midrule
    Omnivore \cite{girdhar2022omnivore} & 47.8 & 62.8 & 35.9 & 58.4 & 86.2 \\
    Student & \textbf{51.7} & \textbf{65.4} & \textbf{39.3} & \textbf{63.0} & \textbf{88.9} \\ \midrule
    \cellcolor{white}{} & \multicolumn{3}{c}{Epic-Kitchens Unseen Participants} & \multicolumn{2}{c}{Something-Else} \\ \midrule
    Omnivore \cite{girdhar2022omnivore} & 38.5 & \textbf{54.5} & 27.9 & 58.3 & 84.9 \\
    Student & \textbf{43.7} & 54.1 & \textbf{29.6} & \textbf{59.1} & \textbf{86.1} \\ \bottomrule
\end{tabular}
}
\caption{Comparison with Omnivorous models \cite{girdhar2022omnivore}. All models are Swin-T and perform inference using only RGB frames. Multimodal distillation using all modalities with $\lambda = 1.0$ and $\gamma = 30.0$.}
\label{table:omnivorous-models}
\end{table}

\subsubsection{Effect on Model Calibration}
\begin{figure}[t]
\centering
\includegraphics[width=0.242\textwidth]{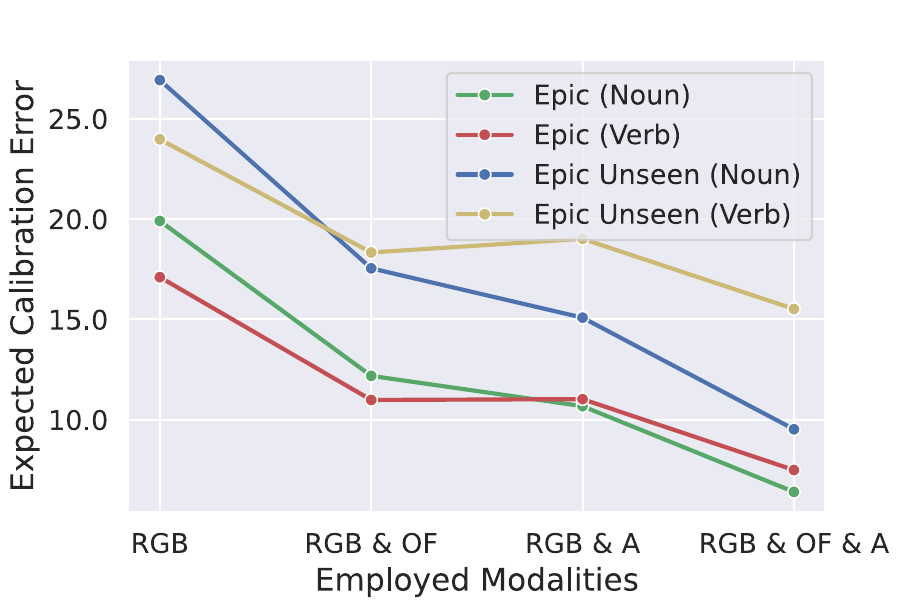}
\hspace{-8pt}
\includegraphics[width=0.242\textwidth]{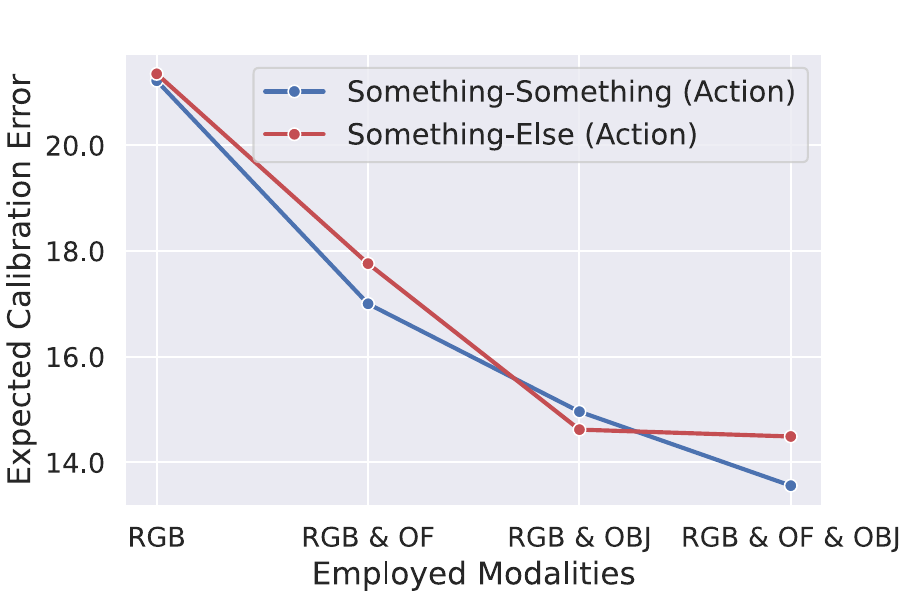}
\caption{Expected Calibration Error across datasets. The student is Swin-T trained with $\lambda = 1.0$ and $\gamma = 30.0$.}
\label{fig:calibration-error}
\end{figure}

Next to the performance improvements we observe, we also investigate how the distilled student fares against the baseline in terms of model calibration. That is, whether the predicted class score reflects the accuracy of predicting the said class \cite{guo2017calibration, popordanoska2022consistent}. More importantly, we verify whether distilling from additional modalities yields better calibrated students. In Fig.~\ref{fig:calibration-error}, we measure the Expected Calibration error \cite{guo2017calibration} (ECE) on the Epic-Kitchens (regular \& unseen) and the Something-Something/Else datasets. We report the ECE for an RGB model trained using the ground truth labels, as well as all distilled students reported in Table~\ref{table:regular-actions} and Table \ref{table:unseen-actions}. The general observation is that \textit{across datasets, distillation improves the models' calibration}. Furthermore, increasing the number of modalities used in the ensemble was found to improve the model calibration.

\subsubsection{Per-Class Performance Breakdown}
\begin{figure}[t]
\centering
\includegraphics[align=t, width=0.49\columnwidth]{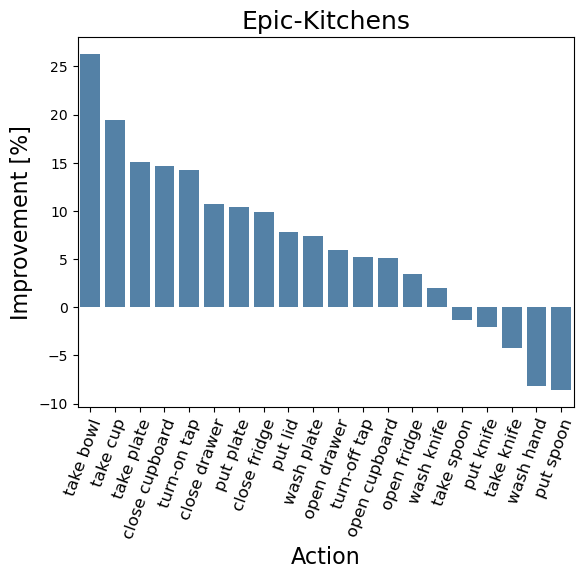}
\includegraphics[align=t, width=0.49\columnwidth]{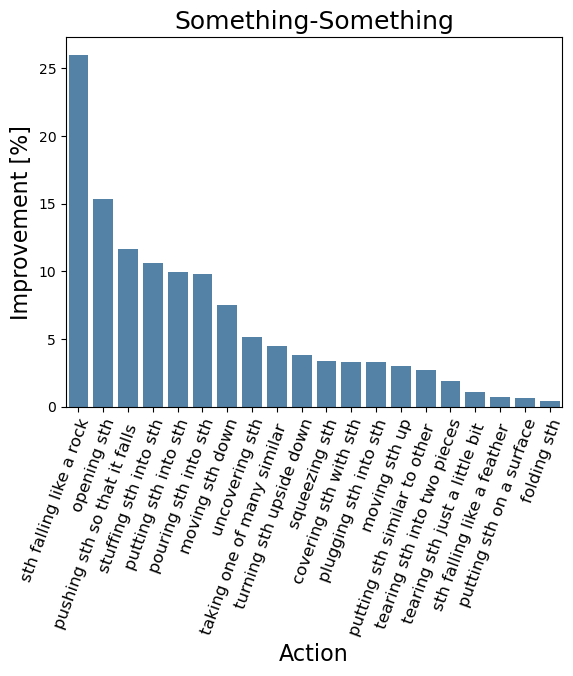}
\caption{Per-action improvement over the RGB baseline on the top-20 most frequent actions across datasets.}
\label{fig:per-class}
\end{figure}

 In Fig~\ref{fig:per-class}, we present the relative change in action recognition accuracy of the student model obtained via multimodal distillation w.r.t.\  the architecturally equivalent baseline RGB model trained on ground truth labels, computed on the top-20 most frequent classes (actions) on Epic-Kitchens and Something-Something. Overall, we observe that multimodal distillation generally improves performance across different actions, and particularly so on Something-Something, where we achieve improvements in terms of all of the top-20 most frequent action classes.

\subsection{Ensemble Teacher Weighting}\label{sec:weighting}
In \S\ref{sec:general-results} we observe that distilling from multimodal teachers yields students which are superior to models trained on the ground truth labels. Nevertheless, if a weak teacher is added in the ensemble, it negatively affects the teaching process and yields a student which is inferior than using the ground truth labels, e.g.\ adding the Audio teacher in the ensemble for Epic-Kitchens. To cope with this, we weigh the logits of the teacher ensemble as discussed in \S\ref{sec:mmkd}. Namely, we use the two hyperparameters ($\lambda$ and $\gamma$) for
\begin{inparaenum}[(i)]
    \item balancing between the ground truth and the distillation loss ($\lambda$), and
    \item controlling how the predictions of modality-specific models are combined in the ensemble ($\gamma$).
\end{inparaenum}
We report results for Epic-Kitchens in Table~\ref{table:ablation-weighting}, including the values for $\lambda$ and $\gamma$ as well as the objective we effectively minimize. We observe that the model trained only on the ground truth labels ($\lambda = 0.0$), is inferior to all other models. Using a large $\gamma$ (e.g.\ $30.0$) effectively assigns equal weights to all models in the ensemble, which we observe to perform well despite the simple setup (used in the experiments in \S\ref{sec:general-results}). Additionally, training using the task loss in addition to the distillation loss ($\lambda = 0.8$) further improves the performance. The best performing model uses both the task and the distillation loss ($\lambda = 0.8$), and assigns weights to each teacher in the ensemble based on its performance on $Z = 1000$ randomly sampled videos held-out from the training dataset, with the normalization temperature $\gamma = 1.0$.
% We do not perform any smoothing ($\gamma > 1.0$) or sharpening ($\gamma < 1.0$) of the teacher weights during distillation ($\gamma = 1.0$), meaning that uncompetitive teachers would get downweighted, e.g., the Audio model for recognizing active objects (nouns).
We find that lowering the normalization temperature $\gamma$ further ($\gamma = 0.33$) -- thus giving higher weight to the best performing modality-specific model in the ensemble -- gives lower performance. Similarly, increasing the normalization temperature to $\gamma = 3.0$ -- such as to equalize the model weights -- also negatively affects the performance.

\begin{table}[t]
\centering
\resizebox{0.65\columnwidth}{!}{
\begin{tabular}{>{\columncolor{graygray}}c>{\columncolor{pinkpink}}c>{\columncolor{yellowyellow}}c>{\columncolor{greengreen}}c>{\columncolor{greengreen}}c>{\columncolor{greengreen}}c} \toprule
    {Objective} & {$\lambda$} & {$\gamma$} & {Noun} & {Verb} & {Action} \\ \midrule
    $\mathcal{L}_{\text{CE}}$ & 0.0 & \NA & 52.0 & 61.7 & 38.3 \\
    $\mathcal{L}_{\text{KL}}$ & 1.0 & 30.0 & 51.7 & 65.4 & 39.3 \\
    $\mathcal{L}_{\text{CE}} \wedge \mathcal{L}_{\text{KL}}$  & 0.8 & 30.0 & 52.6 & 65.1 & 40.4 \\
    $\mathcal{L}_{\text{CE}} \wedge \mathcal{L}_{\text{KL}}$ & 0.8 & 3.0 & 53.0 & 66.9 & 41.0 \\
    $\mathcal{L}_{\text{KL}}$ & 1.0 & 1.0 & 53.1 & 65.5 & 40.5 \\
    $\mathcal{L}_{\text{CE}} \wedge \mathcal{L}_{\text{KL}}$ & 0.8 & 1.0 & 53.5 & 65.4 & 41.2 \\
    $\mathcal{L}_{\text{CE}} \wedge \mathcal{L}_{\text{KL}}$ & 0.8 & 0.33 & 53.6 & 64.7 & 40.5 \\ \bottomrule
\end{tabular}
}
\caption{Ablation study on the Epic-Kitchens dataset. $\lambda$: Distillation and Cross-Entropy loss balancing term; $\gamma$: Temperature of the Ensemble Teacher Weighting.}
\label{table:ablation-weighting}
\end{table}

\subsection{Efficiency Analysis}\label{sec:speed}
The teacher ensemble used for Epic-Kitchens uses three modality-specific Swin-T models, where each has 28.22M parameters, and requires 140.33 GFLOPs for processing single-view 32 frame/spectrogram video. Assuming such an ensemble is deployed, the optical flow would have to be computed on-the-fly\footnote{The Duality-based TV-L1 \cite{zach2007duality, perez2013tv} can be efficiently computed on a GPU (5-10 FPS) \cite{bao2014comparison}. Deep Learning-based approaches, e.g.\ RAFT \cite{teed2020raft}, require 163.37 GFLOPs, however, achieve higher FPS of 21.10, measured with 10 refinement iterations and resolution of $256 \times 456$.}. Using RAFT, we measure a total added computation of 163.37 GFLOPs for such a model. We consider the computation required to obtain the spectrograms of 1.116s audio segments to be negligible in comparison. Therefore, when using all three modalities, updating the input sequences for each newly observed frame and performing action recognition would require 584.36 GFLOPs. In contrast, the distilled student is a single RGB model, and in the case of Swin-T requires 140.33 GFLOPs -- \textit{which represents a reduction of 75.98\%.}\par
\begin{figure*}[t]
\centering
\resizebox{0.83\textwidth}{!}{
\includegraphics[width=1.0\columnwidth]{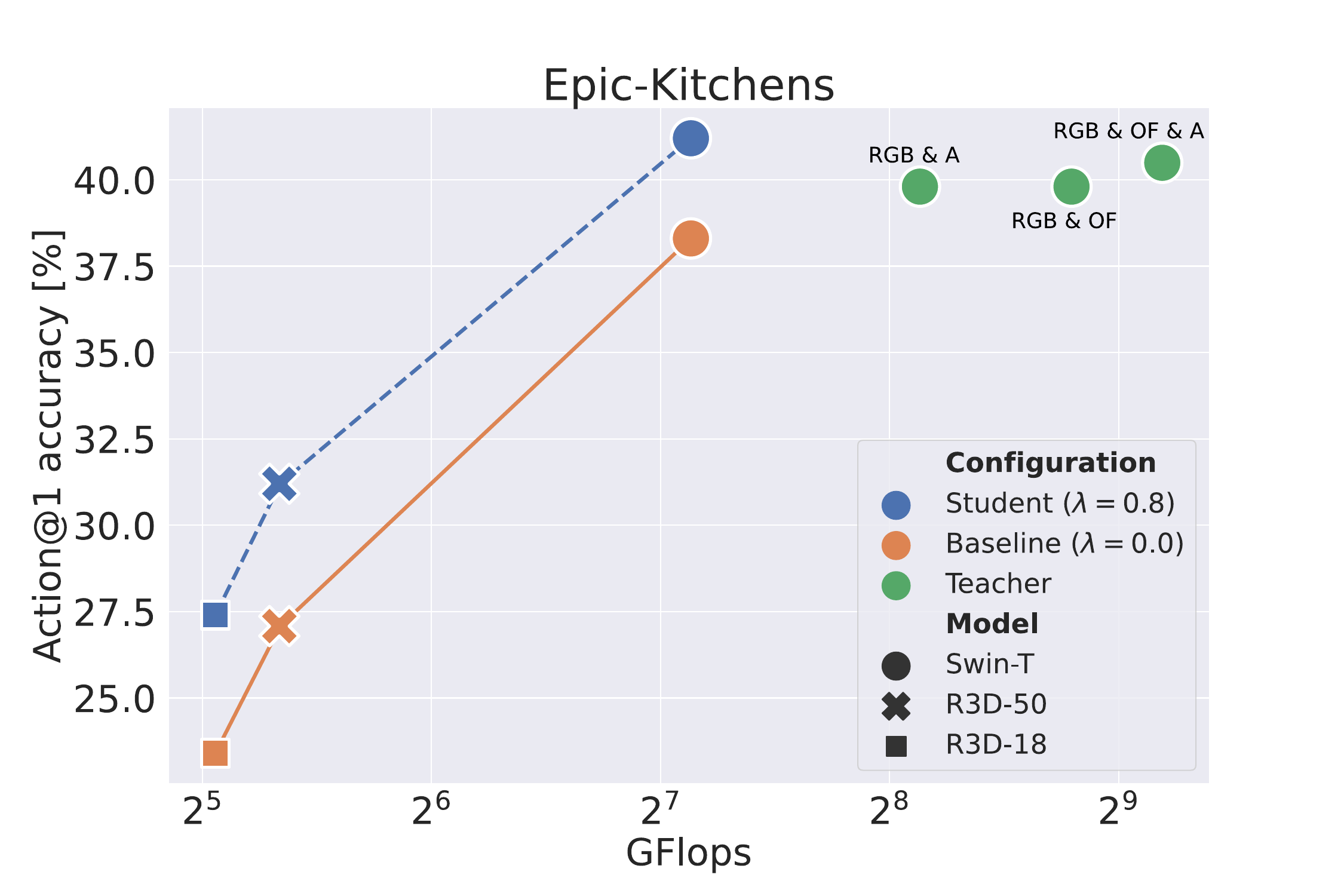}
\includegraphics[width=1.0\columnwidth]{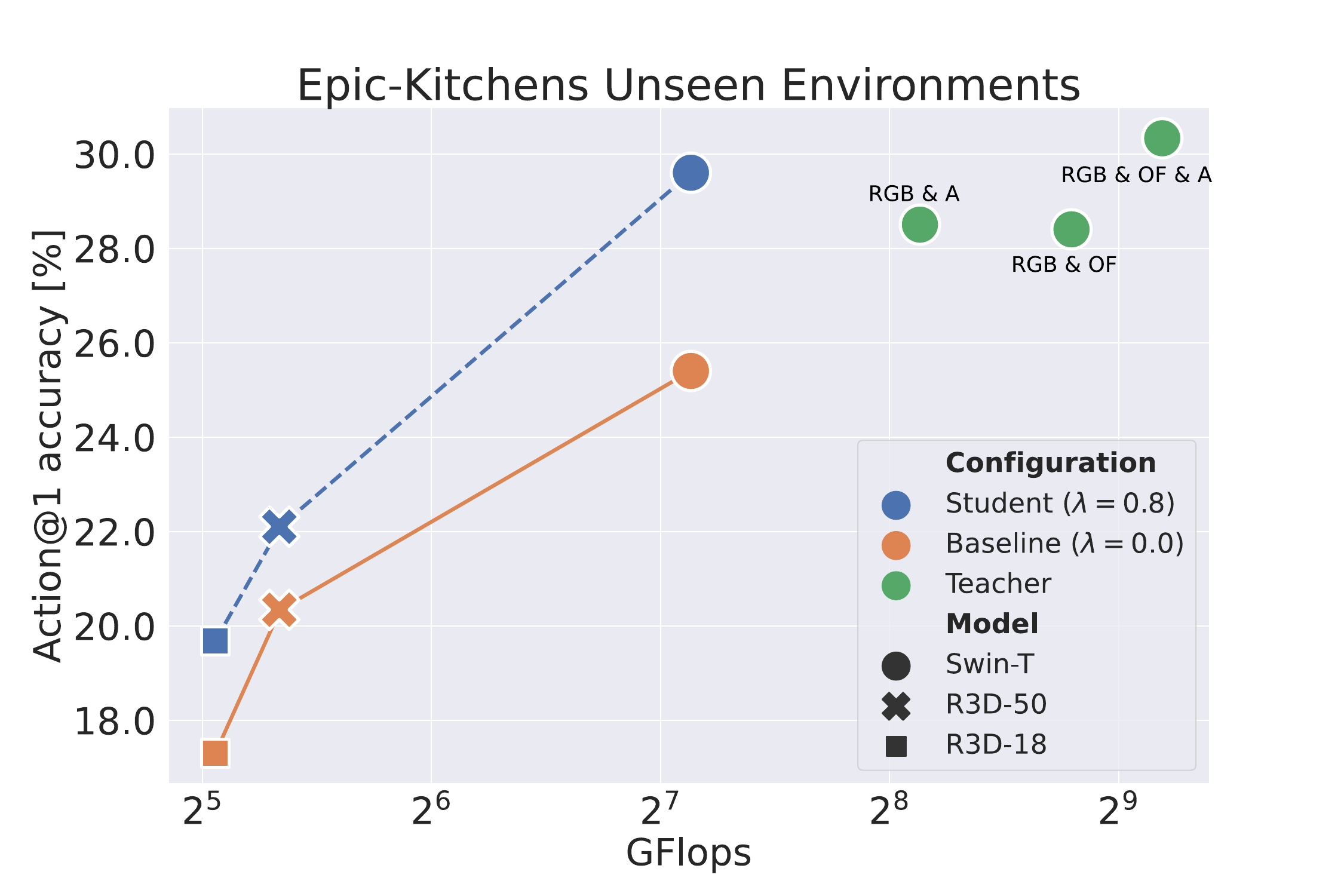}
% \includegraphics[width=0.265\textwidth]{images/something-something-models.pdf}
% \hspace{-17pt}
% \includegraphics[width=0.265\textwidth]{images/something-else-models.pdf}
}
\caption{Top-1 action accuracy of \textcolor{teacher}{Teacher}, \textcolor{student}{Student} \& \textcolor{baseline}{Baseline} models and their associated computational cost in giga-FLOPs ($10^9$) required to update the input and predict the action. Note: Top-left corner is optimal (faster \& most accurate).}
\label{fig:model-families-complexity}
\end{figure*}

% \begin{itemize}
%     \item V+A = 280.66
%     \item V+OF = 444.03
%     \item V+A+OF = 584.36
% \end{itemize}
% We measure throughput on a single machine machine (Intel(R) Core(TM) i7-4770K CPU @ 3.50GHz and an NVIDIA TITAN Xp) as the average number of videos processed per second over 100 batches, using the largest possible batch size for each model.
% This amounts to 84.66M parameters and 420.99 GFLOPs for the input modalities.
We report results on the Epic-Kitchens dataset in Fig.~\ref{fig:model-families-complexity} for:
\begin{inparaenum}[(i)]
    \item All variations of teacher models (RGB \& OF, RGB \& A, and RGB \& OF \& A);
    \item The Swin-T student model, distilled with $\lambda = 0.8$ and $\gamma = 1.0$;
    \item The Swin-T baseline model, trained with $\lambda = 0.0$;
    \item Two new ResNet3D models \cite{kataoka2020would} (depth 50 and 18), exhibiting less parameters and GFLOPs compared to Swin-T. The resolution size of the ResNet3D models is $112 \times 112$. For each R3D, we report baselines with $\lambda = 0.0$, and distilled students with $\lambda = 0.8$ and $\gamma = 1.0$.
\end{inparaenum}\par
We observe a consistently higher performance of the student models than their counterparts trained on ground truth labels alone. Notably, our best performing student achieves comparable performance to the significantly more expensive RGB \& OF \& A teacher. Furthermore, the R3D-18 and R3D-50 students outperform their counterparts trained on class labels. Finally, we observe that the R3D-18 student matches the performance of the larger, and computationally more expensive R3D-50 trained on ground truth labels.

\subsubsection{Effect of Test-Time Augmentation}
\begin{figure}[t]
\centering
\includegraphics[width=0.25\textwidth]{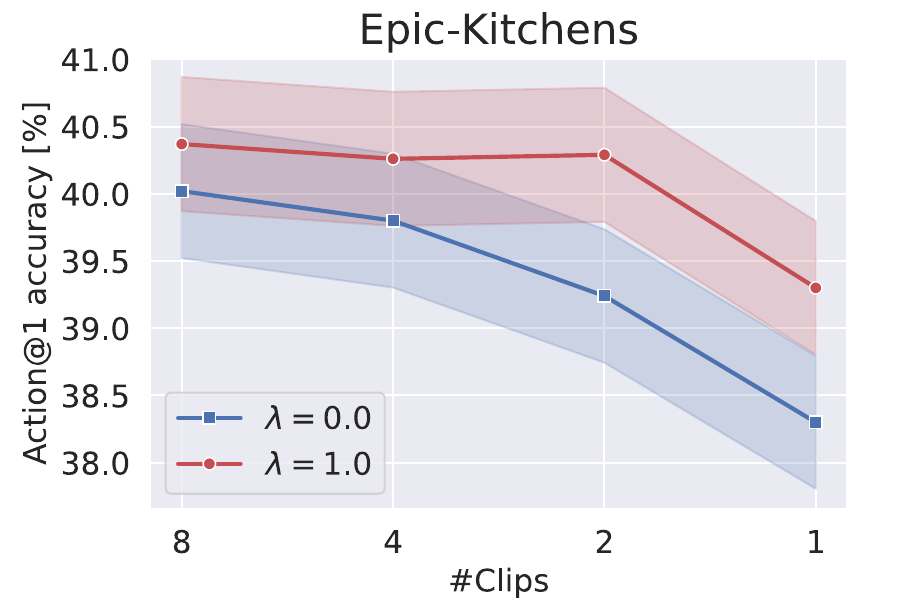}
\hspace{-16pt}
\includegraphics[width=0.25\textwidth]{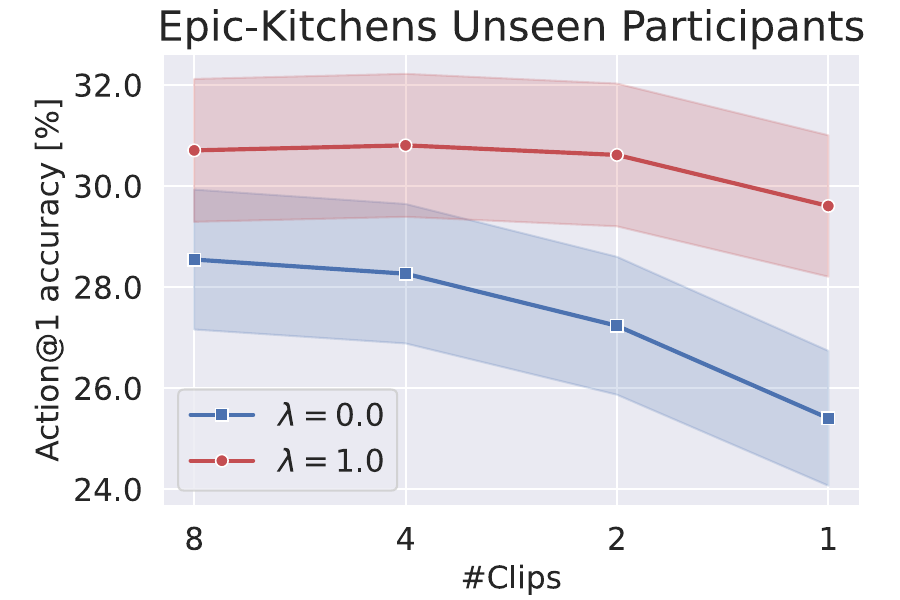}
\hspace{-16pt}
\includegraphics[width=0.25\textwidth]{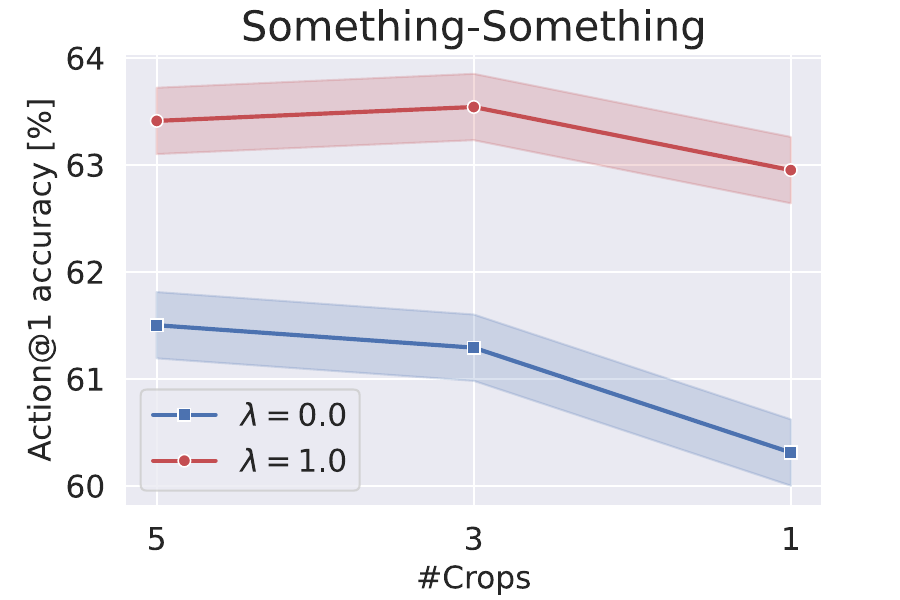}
\hspace{-16pt}
\includegraphics[width=0.25\textwidth]{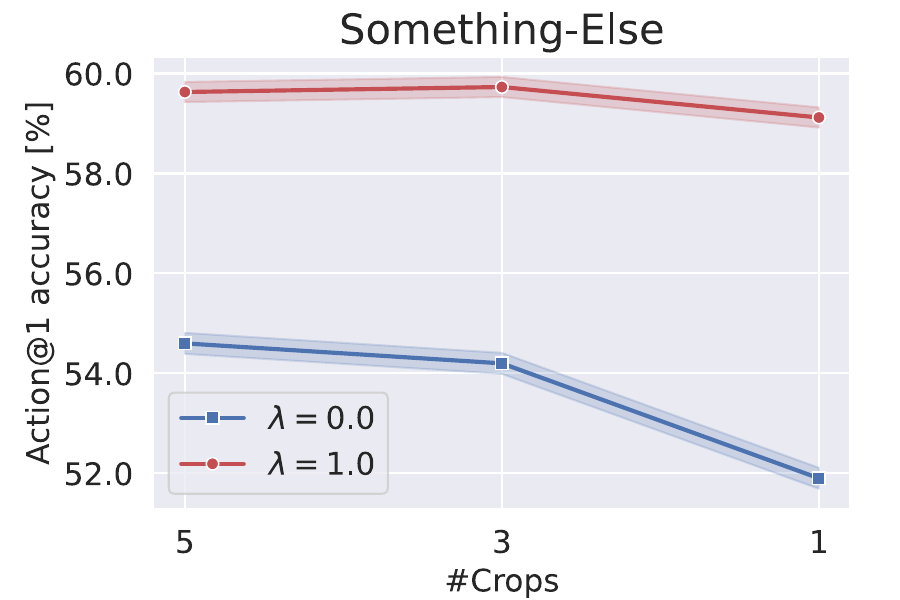}
\caption{Performance degradation when reducing the number of inference clips/crops (Epic-Kitchens/Something-Something).}
\label{fig:ablation_num_clips}
\end{figure}

Lastly, we inspect the relationship between performance and the number of temporal clips (on Epic-Kitchens) and spatial crops (on Something-Something) used during inference. Our goal is to reduce the computational complexity while maintaining the performance. Since standard video models \cite{bulat2021space, liu2021swin, li2022mvitv2, arnab2021vivit} use multiple temporal clips and spatial crops as test-time augmentation, we explore the extent to which the distilled model is dependent on their availability. We report results in Fig.~\ref{fig:ablation_num_clips}, where we observe that the distilled model ($\lambda = 1.0$) is much less adversely affected by the reduction of both sampled temporal clips and spatial crops during inference, compared to the same model trained on the ground truth labels ($\lambda = 0.0$).

\subsection{Qualitative Examples}
\begin{figure}[t]
\centering
\includegraphics[width=\columnwidth]{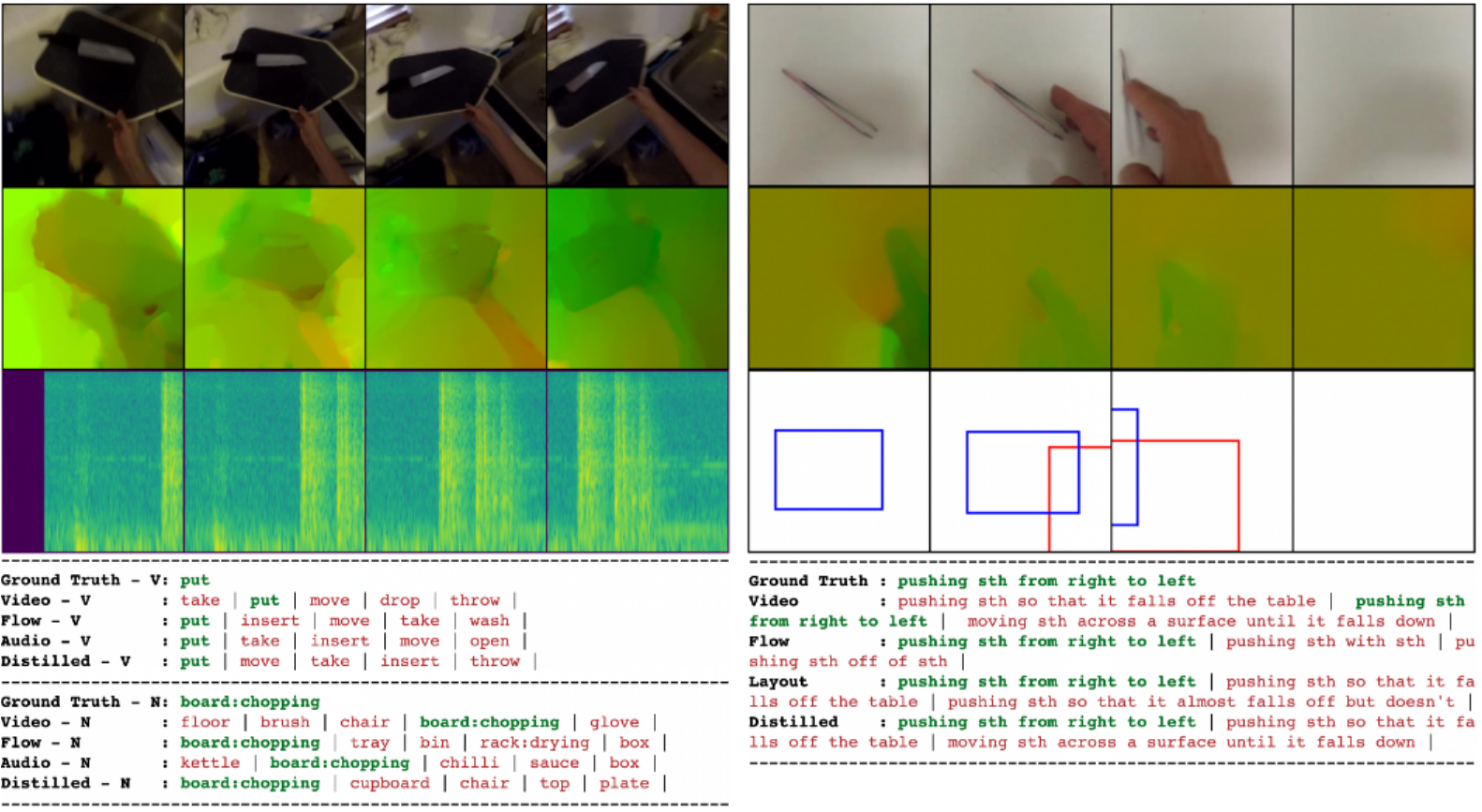}
\caption{Qualitative example for the Epic Kitchens (Left) \& the Something-Something (Right) datasets.}
\label{fig:qualitative}
\end{figure}

In Fig.~\ref{fig:qualitative}, we showcase the classes corresponding to the highest scores predicted by the student and the individual modality-specific models in the teacher ensemble, as well as the ground truth label (on Epic-Kitchens and Something-Something datasets). For both examples, we observe that the distilled student picks up on relevant cues from each modality and accurately predicts the action of interest (see Supp. for additional qualitative examples).

% We conduct an experiment to evaluate the robustness of our approach to the reduction of the number of frames extracted from each action sequence during training and inference on the Something-Else compositional split of the Something-Something dataset. In the case of the distilled model, we use the multimodal teacher trained on 16-frame action sequences, and perform knowledge distillation to the RGB student using 8, 4, 2 and 1 frame, respectively. We compare the robustness of our distilled model with the RGB student using the same numbers of frames, but distilled from a single, RGB teacher using 16 frames. As can be seen on Figure~XX, distilling from a multi-modal teacher leads to much less performance degradation when a lower number of frames is used for the student, than when distiling from a single modality.

% \begin{figure}[t]
% \centering
% \includegraphics[width=0.3\textwidth]{images/something_else_frames.pdf}
% \caption{Robust-ness to performance degradation due to reduction of the number of video frames used during training and inference.}
% \label{fig:ablation_num_frames}
% \end{figure}

% \begin{figure*}[t]
% \begin{center}
%  \centering
% \subfloat[Epic-Kitchens 100]{\includegraphics[width=0.9\linewidth]{images/EpicRegularNumClips.pdf}}
% \vfill
% \subfloat[Epic-Kitchens 100 Unseen Split]{\includegraphics[width=0.9\linewidth]{images/EpicUnseenNumClips.pdf}}
% \end{center}
% \caption{Robus-tness to performance degradation due to reduction of the number of video clips used during inference.}
% \label{fig:ablation_num_clips}
% \end{figure*}
\section{Conclusion}\label{sec:conclusion}
We demonstrated a simple, yet effective distillation-based approach for leveraging multimodal data \textit{only} during training in order to improve a model that uses \textit{solely} RGB frames during inference. Our experiments indicate clear performance improvements with respect to models trained on ground truth labels. We further showed an advantageous trade-off between the high performance of a cumbersome multimodal ensemble, and low computational complexity of unimodal approaches. Moreover, our approach relies less on expensive test-time augmentations, otherwise widely used in literature to improve the egocentric action recognition models' performance.\par
\textbf{Future work \& Limitations.} Notably, in this work we considered only the task of action recognition, while multimodal distillation can be readily applied to other egocentric tasks \cite{grauman2022ego4d}. Furthermore, we considered a limited set of modalities and future work may include additional modalities such as depth, hand poses, motion captured by inertial sensors (IMU), etc.
% We demonstrate a simple, yet effective distillation-based approach for leveraging multiple data modalities at train time for improving the performance of a light weight model which uses only RGB video as input modality during inference. Our experiments show clear performance improvements with respect to baseline models trained on ground truth labels. We further show that we achieve an advantageous trade-off by which maintain comparable performance to the cumbersome multimodal teacher ensemble, while achieving the low computational complexity of the unimodal approaches. Our approach outperforms the Omnivore variant of our model, where an architecturally equivalent model was trained while interchangeably using data from different modalities. 

% Future work may extend our approach to the larger, EGO4D dataset, featuring additional data and diverse task setups. Furthermore, our approach can be seamlessly extended to distilling multi-view models onto models that use egocentric video alone. 

\section*{Acknowledgement}\label{sec:aknowledgement}
We acknowledge funding from the Flemish Government under the Onderzoeksprogramma Artifici\"{e}le Intelligentie (AI) Vlaanderen programme.

\nocite{radevski2022students}
{\small
\bibliographystyle{ieee_fullname}
\bibliography{egbib}

\begin{thebibliography}{10}\itemsep=-1pt

\bibitem{arnab2021vivit}
Anurag Arnab, Mostafa Dehghani, Georg Heigold, Chen Sun, Mario Lu{\v{c}}i{\'c},
  and Cordelia Schmid.
\newblock Vivit: A video vision transformer.
\newblock In {\em Proceedings of the IEEE/CVF International Conference on
  Computer Vision}, pages 6836--6846, 2021.

\bibitem{asif2019ensemble}
Umar Asif, Jianbin Tang, and Stefan Harrer.
\newblock Ensemble knowledge distillation for learning improved and efficient
  networks.
\newblock {\em arXiv preprint arXiv:1909.08097}, 2019.

\bibitem{aytar2016soundnet}
Yusuf Aytar, Carl Vondrick, and Antonio Torralba.
\newblock Soundnet: Learning sound representations from unlabeled video.
\newblock {\em Advances in neural information processing systems}, 29, 2016.

\bibitem{bao2014comparison}
Linchao Bao, Hailin Jin, Byungmoon Kim, and Qingxiong Yang.
\newblock A comparison of tv-l1 optical flow solvers on gpu.
\newblock {\em GTC Posters}, 6, 2014.

\bibitem{bertasius2021space}
Gedas Bertasius, Heng Wang, and Lorenzo Torresani.
\newblock Is space-time attention all you need for video understanding?
\newblock In {\em ICML}, volume~2, page~4, 2021.

\bibitem{beyer2022knowledge}
Lucas Beyer, Xiaohua Zhai, Am{\'e}lie Royer, Larisa Markeeva, Rohan Anil, and
  Alexander Kolesnikov.
\newblock Knowledge distillation: A good teacher is patient and consistent.
\newblock In {\em Proceedings of the IEEE/CVF Conference on Computer Vision and
  Pattern Recognition}, pages 10925--10934, 2022.

\bibitem{bulat2021space}
Adrian Bulat, Juan~Manuel Perez~Rua, Swathikiran Sudhakaran, Brais Martinez,
  and Georgios Tzimiropoulos.
\newblock Space-time mixing attention for video transformer.
\newblock {\em Advances in Neural Information Processing Systems},
  34:19594--19607, 2021.

\bibitem{carreira2017quo}
Joao Carreira and Andrew Zisserman.
\newblock Quo vadis, action recognition? a new model and the kinetics dataset.
\newblock In {\em proceedings of the IEEE Conference on Computer Vision and
  Pattern Recognition}, pages 6299--6308, 2017.

\bibitem{chen2022dearkd}
Xianing Chen, Qiong Cao, Yujie Zhong, Jing Zhang, Shenghua Gao, and Dacheng
  Tao.
\newblock Dearkd: Data-efficient early knowledge distillation for vision
  transformers.
\newblock In {\em Proceedings of the IEEE/CVF Conference on Computer Vision and
  Pattern Recognition}, pages 12052--12062, 2022.

\bibitem{cho2019efficacy}
Jang~Hyun Cho and Bharath Hariharan.
\newblock On the efficacy of knowledge distillation.
\newblock In {\em Proceedings of the IEEE/CVF international conference on
  computer vision}, pages 4794--4802, 2019.

\bibitem{dai2022one}
Yong Dai, Duyu Tang, Liangxin Liu, Minghuan Tan, Cong Zhou, Jingquan Wang,
  Zhangyin Feng, Fan Zhang, Xueyu Hu, and Shuming Shi.
\newblock One model, multiple modalities: A sparsely activated approach for
  text, sound, image, video and code.
\newblock {\em arXiv preprint arXiv:2205.06126}, 2022.

\bibitem{damen2018scaling}
Dima Damen, Hazel Doughty, Giovanni~Maria Farinella, Sanja Fidler, Antonino
  Furnari, Evangelos Kazakos, Davide Moltisanti, Jonathan Munro, Toby Perrett,
  Will Price, et~al.
\newblock Scaling egocentric vision: The epic-kitchens dataset.
\newblock In {\em Proceedings of the European Conference on Computer Vision
  (ECCV)}, pages 720--736, 2018.

\bibitem{damen2020rescaling}
Dima Damen, Hazel Doughty, Giovanni~Maria Farinella, Antonino Furnari,
  Evangelos Kazakos, Jian Ma, Davide Moltisanti, Jonathan Munro, Toby Perrett,
  Will Price, et~al.
\newblock Rescaling egocentric vision.
\newblock {\em arXiv preprint arXiv:2006.13256}, 2020.

\bibitem{furnari2019would}
Antonino Furnari and Giovanni~Maria Farinella.
\newblock What would you expect? anticipating egocentric actions with
  rolling-unrolling lstms and modality attention.
\newblock In {\em Proceedings of the IEEE/CVF International Conference on
  Computer Vision}, pages 6252--6261, 2019.

\bibitem{gabeur2020multi}
Valentin Gabeur, Chen Sun, Karteek Alahari, and Cordelia Schmid.
\newblock Multi-modal transformer for video retrieval.
\newblock In {\em European Conference on Computer Vision}, pages 214--229.
  Springer, 2020.

\bibitem{garcia2019dmcl}
Nuno~C Garcia, Sarah~Adel Bargal, Vitaly Ablavsky, Pietro Morerio, Vittorio
  Murino, and Stan Sclaroff.
\newblock Dmcl: Distillation multiple choice learning for multimodal action
  recognition.
\newblock {\em arXiv preprint arXiv:1912.10982}, 2019.

\bibitem{garcia2018modality}
Nuno~C Garcia, Pietro Morerio, and Vittorio Murino.
\newblock Modality distillation with multiple stream networks for action
  recognition.
\newblock In {\em Proceedings of the European Conference on Computer Vision
  (ECCV)}, pages 103--118, 2018.

\bibitem{girdhar2022omnimae}
Rohit Girdhar, Alaaeldin El-Nouby, Mannat Singh, Kalyan~Vasudev Alwala, Armand
  Joulin, and Ishan Misra.
\newblock Omnimae: Single model masked pretraining on images and videos.
\newblock {\em arXiv preprint arXiv:2206.08356}, 2022.

\bibitem{girdhar2022omnivore}
Rohit Girdhar, Mannat Singh, Nikhila Ravi, Laurens van~der Maaten, Armand
  Joulin, and Ishan Misra.
\newblock Omnivore: A single model for many visual modalities.
\newblock In {\em Proceedings of the IEEE/CVF Conference on Computer Vision and
  Pattern Recognition}, pages 16102--16112, 2022.

\bibitem{goyal2017something}
Raghav Goyal, Samira Ebrahimi~Kahou, Vincent Michalski, Joanna Materzynska,
  Susanne Westphal, Heuna Kim, Valentin Haenel, Ingo Fruend, Peter Yianilos,
  Moritz Mueller-Freitag, et~al.
\newblock The" something something" video database for learning and evaluating
  visual common sense.
\newblock In {\em Proceedings of the IEEE international conference on computer
  vision}, pages 5842--5850, 2017.

\bibitem{grauman2022ego4d}
Kristen Grauman, Andrew Westbury, Eugene Byrne, Zachary Chavis, Antonino
  Furnari, Rohit Girdhar, Jackson Hamburger, Hao Jiang, Miao Liu, Xingyu Liu,
  et~al.
\newblock Ego4d: Around the world in 3,000 hours of egocentric video.
\newblock In {\em Proceedings of the IEEE/CVF Conference on Computer Vision and
  Pattern Recognition}, pages 18995--19012, 2022.

\bibitem{guo2017calibration}
Chuan Guo, Geoff Pleiss, Yu Sun, and Kilian~Q Weinberger.
\newblock On calibration of modern neural networks.
\newblock In {\em International conference on machine learning}, pages
  1321--1330. PMLR, 2017.

\bibitem{gupta2016cross}
Saurabh Gupta, Judy Hoffman, and Jitendra Malik.
\newblock Cross modal distillation for supervision transfer.
\newblock In {\em Proceedings of the IEEE conference on computer vision and
  pattern recognition}, pages 2827--2836, 2016.

\bibitem{herzig2022object}
Roei Herzig, Elad Ben-Avraham, Karttikeya Mangalam, Amir Bar, Gal Chechik, Anna
  Rohrbach, Trevor Darrell, and Amir Globerson.
\newblock Object-region video transformers.
\newblock In {\em Proceedings of the IEEE/CVF Conference on Computer Vision and
  Pattern Recognition}, pages 3148--3159, 2022.

\bibitem{hinton2015distilling}
Geoffrey Hinton, Oriol Vinyals, Jeff Dean, et~al.
\newblock Distilling the knowledge in a neural network.
\newblock {\em arXiv preprint arXiv:1503.02531}, 2(7), 2015.

\bibitem{kataoka2020would}
Hirokatsu Kataoka, Tenga Wakamiya, Kensho Hara, and Yutaka Satoh.
\newblock Would mega-scale datasets further enhance spatiotemporal 3d cnns?
\newblock {\em arXiv preprint arXiv:2004.04968}, 2020.

\bibitem{kazakos2021little}
Evangelos Kazakos, Jaesung Huh, Arsha Nagrani, Andrew Zisserman, and Dima
  Damen.
\newblock With a little help from my temporal context: Multimodal egocentric
  action recognition.
\newblock {\em arXiv preprint arXiv:2111.01024}, 2021.

\bibitem{kim2021motion}
Tae~Soo Kim, Jonathan Jones, and Gregory~D Hager.
\newblock Motion guided attention fusion to recognize interactions from videos.
\newblock In {\em Proceedings of the IEEE/CVF International Conference on
  Computer Vision}, pages 13076--13086, 2021.

\bibitem{li2022mvitv2}
Yanghao Li, Chao-Yuan Wu, Haoqi Fan, Karttikeya Mangalam, Bo Xiong, Jitendra
  Malik, and Christoph Feichtenhofer.
\newblock Mvitv2: Improved multiscale vision transformers for classification
  and detection.
\newblock In {\em Proceedings of the IEEE/CVF Conference on Computer Vision and
  Pattern Recognition}, pages 4804--4814, 2022.

\bibitem{liu2021swin}
Ze Liu, Yutong Lin, Yue Cao, Han Hu, Yixuan Wei, Zheng Zhang, Stephen Lin, and
  Baining Guo.
\newblock Swin transformer: Hierarchical vision transformer using shifted
  windows.
\newblock In {\em Proceedings of the IEEE/CVF International Conference on
  Computer Vision}, pages 10012--10022, 2021.

\bibitem{loshchilov2017decoupled}
Ilya Loshchilov and Frank Hutter.
\newblock Decoupled weight decay regularization.
\newblock {\em arXiv preprint arXiv:1711.05101}, 2017.

\bibitem{materzynska2020something}
Joanna Materzynska, Tete Xiao, Roei Herzig, Huijuan Xu, Xiaolong Wang, and
  Trevor Darrell.
\newblock Something-else: Compositional action recognition with
  spatial-temporal interaction networks.
\newblock In {\em Proceedings of the IEEE/CVF Conference on Computer Vision and
  Pattern Recognition}, pages 1049--1059, 2020.

\bibitem{menon2021statistical}
Aditya~K Menon, Ankit~Singh Rawat, Sashank Reddi, Seungyeon Kim, and Sanjiv
  Kumar.
\newblock A statistical perspective on distillation.
\newblock In {\em International Conference on Machine Learning}, pages
  7632--7642. PMLR, 2021.

\bibitem{mishra2017apprentice}
Asit Mishra and Debbie Marr.
\newblock Apprentice: Using knowledge distillation techniques to improve
  low-precision network accuracy.
\newblock {\em arXiv preprint arXiv:1711.05852}, 2017.

\bibitem{nagrani2021attention}
Arsha Nagrani, Shan Yang, Anurag Arnab, Aren Jansen, Cordelia Schmid, and Chen
  Sun.
\newblock Attention bottlenecks for multimodal fusion.
\newblock {\em Advances in Neural Information Processing Systems},
  34:14200--14213, 2021.

\bibitem{neverova2015moddrop}
Natalia Neverova, Christian Wolf, Graham Taylor, and Florian Nebout.
\newblock Moddrop: adaptive multi-modal gesture recognition.
\newblock {\em IEEE Transactions on Pattern Analysis and Machine Intelligence},
  38(8):1692--1706, 2015.

\bibitem{park2019specaugment}
Daniel~S Park, William Chan, Yu Zhang, Chung-Cheng Chiu, Barret Zoph, Ekin~D
  Cubuk, and Quoc~V Le.
\newblock Specaugment: A simple data augmentation method for automatic speech
  recognition.
\newblock {\em arXiv preprint arXiv:1904.08779}, 2019.

\bibitem{parthasarathy2020training}
Srinivas Parthasarathy and Shiva Sundaram.
\newblock Training strategies to handle missing modalities for audio-visual
  expression recognition.
\newblock In {\em Companion Publication of the 2020 International Conference on
  Multimodal Interaction}, pages 400--404, 2020.

\bibitem{perez2013tv}
Javier~S{\'a}nchez P{\'e}rez, Enric Meinhardt-Llopis, and Gabriele Facciolo.
\newblock Tv-l1 optical flow estimation.
\newblock {\em Image Processing On Line}, 2013:137--150, 2013.

\bibitem{popordanoska2022consistent}
Teodora Popordanoska, Raphael Sayer, and Matthew~B Blaschko.
\newblock A consistent and differentiable $l_p$ canonical calibration error
  estimator.
\newblock In {\em Advances in Neural Information Processing Systems}, 2022.

\bibitem{radevski2022students}
Gorjan Radevski, Dusan Grujicic, Matthew Blaschko, Marie-Francine Moens, and
  Tinne Tuytelaars.
\newblock Students taught by multimodal teachers are superior action
  recognizers.
\newblock {\em arXiv preprint arXiv:2210.04331}, 2022.

\bibitem{radevski2021revisiting}
Gorjan Radevski, Marie-Francine Moens, and Tinne Tuytelaars.
\newblock Revisiting spatio-temporal layouts for compositional action
  recognition.
\newblock {\em arXiv preprint arXiv:2111.01936}, 2021.

\bibitem{ren2015faster}
Shaoqing Ren, Kaiming He, Ross Girshick, and Jian Sun.
\newblock Faster r-cnn: Towards real-time object detection with region proposal
  networks.
\newblock {\em Advances in neural information processing systems}, 28, 2015.

\bibitem{rousseau2021post}
Axel-Jan Rousseau, Thijs Becker, Jeroen Bertels, Matthew~B Blaschko, and Dirk
  Valkenborg.
\newblock Post training uncertainty calibration of deep networks for medical
  image segmentation.
\newblock In {\em 2021 IEEE 18th International Symposium on Biomedical Imaging
  (ISBI)}, pages 1052--1056. IEEE, 2021.

\bibitem{shan2020understanding}
Dandan Shan, Jiaqi Geng, Michelle Shu, and David~F Fouhey.
\newblock Understanding human hands in contact at internet scale.
\newblock In {\em Proceedings of the IEEE/CVF conference on computer vision and
  pattern recognition}, pages 9869--9878, 2020.

\bibitem{shen2020meal}
Zhiqiang Shen and Marios Savvides.
\newblock Meal v2: Boosting vanilla resnet-50 to 80\%+ top-1 accuracy on
  imagenet without tricks.
\newblock {\em arXiv preprint arXiv:2009.08453}, 2020.

\bibitem{stergiou2022play}
Alexandros Stergiou and Dima Damen.
\newblock Play it back: Iterative attention for audio recognition.
\newblock {\em arXiv preprint arXiv:2210.11328}, 2022.

\bibitem{tan2023egodistill}
Shuhan Tan, Tushar Nagarajan, and Kristen Grauman.
\newblock Egodistill: Egocentric head motion distillation for efficient video
  understanding.
\newblock {\em arXiv preprint arXiv:2301.02217}, 2023.

\bibitem{teed2020raft}
Zachary Teed and Jia Deng.
\newblock Raft: Recurrent all-pairs field transforms for optical flow.
\newblock In {\em Computer Vision--ECCV 2020: 16th European Conference,
  Glasgow, UK, August 23--28, 2020, Proceedings, Part II 16}, pages 402--419.
  Springer, 2020.

\bibitem{touvron2021training}
Hugo Touvron, Matthieu Cord, Matthijs Douze, Francisco Massa, Alexandre
  Sablayrolles, and Herv{\'e} J{\'e}gou.
\newblock Training data-efficient image transformers \& distillation through
  attention.
\newblock In {\em International Conference on Machine Learning}, pages
  10347--10357. PMLR, 2021.

\bibitem{vaswani2017attention}
Ashish Vaswani, Noam Shazeer, Niki Parmar, Jakob Uszkoreit, Llion Jones,
  Aidan~N Gomez, {\L}ukasz Kaiser, and Illia Polosukhin.
\newblock Attention is all you need.
\newblock {\em Advances in neural information processing systems}, 30, 2017.

\bibitem{wang2016temporal}
Limin Wang, Yuanjun Xiong, Zhe Wang, Yu Qiao, Dahua Lin, Xiaoou Tang, and
  Luc~Van Gool.
\newblock Temporal segment networks: Towards good practices for deep action
  recognition.
\newblock In {\em European conference on computer vision}, pages 20--36.
  Springer, 2016.

\bibitem{wang2021knowledge}
Lin Wang and Kuk-Jin Yoon.
\newblock Knowledge distillation and student-teacher learning for visual
  intelligence: A review and new outlooks.
\newblock {\em IEEE Transactions on Pattern Analysis and Machine Intelligence},
  2021.

\bibitem{wang2018videos}
Xiaolong Wang and Abhinav Gupta.
\newblock Videos as space-time region graphs.
\newblock In {\em Proceedings of the European conference on computer vision
  (ECCV)}, pages 399--417, 2018.

\bibitem{xiong2022m}
Xuehan Xiong, Anurag Arnab, Arsha Nagrani, and Cordelia Schmid.
\newblock M\&m mix: A multimodal multiview transformer ensemble.
\newblock {\em arXiv preprint arXiv:2206.09852}, 2022.

\bibitem{xue2021multimodal}
Zihui Xue, Sucheng Ren, Zhengqi Gao, and Hang Zhao.
\newblock Multimodal knowledge expansion.
\newblock In {\em Proceedings of the IEEE/CVF International Conference on
  Computer Vision}, pages 854--863, 2021.

\bibitem{yan2020interactive}
Rui Yan, Lingxi Xie, Xiangbo Shu, and Jinhui Tang.
\newblock Interactive fusion of multi-level features for compositional activity
  recognition.
\newblock {\em arXiv preprint arXiv:2012.05689}, 2020.

\bibitem{zach2007duality}
Christopher Zach, Thomas Pock, and Horst Bischof.
\newblock A duality based approach for realtime tv-l 1 optical flow.
\newblock In {\em Joint pattern recognition symposium}, pages 214--223.
  Springer, 2007.

\bibitem{zhang2022object}
Chuhan Zhang, Ankush Gupta, and Andrew Zisserman.
\newblock Is an object-centric video representation beneficial for transfer?
\newblock {\em arXiv preprint arXiv:2207.10075}, 2022.

\bibitem{zhao2021missing}
Jinming Zhao, Ruichen Li, and Qin Jin.
\newblock Missing modality imagination network for emotion recognition with
  uncertain missing modalities.
\newblock In {\em Proceedings of the 59th Annual Meeting of the Association for
  Computational Linguistics and the 11th International Joint Conference on
  Natural Language Processing (Volume 1: Long Papers)}, pages 2608--2618, 2021.

\bibitem{zhou2015temporal}
Yipin Zhou and Tamara~L Berg.
\newblock Temporal perception and prediction in ego-centric video.
\newblock In {\em Proceedings of the IEEE International Conference on Computer
  Vision}, pages 4498--4506, 2015.

\end{thebibliography}
}

\clearpage
\appendix
\section*{Supplementary material}
The Supplementary material is organized as follows:

\begin{itemize}
    \item Extended discussion and details regarding the datasets we use (\S\ref{supp:datasets}).
    \item Additional details about each of the modalities, as well as the modality-specific models we use (\S\ref{supp:modalities}).
    \item An ablation study, alike the one in Table~5 (main paper), conducted on the Something-Something and Something-Else datasets (\S\ref{supp:weighting}).
    % \item Details about models trained using multimodal data, however, perform inference using only RGB frames, including the Omnivore model -- used for the experiments in Section~4.1.3 in the main paper 
    \item Details on approaches which leverage multimodal data during training (\S\ref{supp:omnivore}).
    \item Learning curves on the Epic-Kitchens and Something-Something datasets (\S\ref{supp:learning_curves}).
    \item Extended per-class performance breakdown for Epic-Kitchens and Something-Something (\S\ref{supp:performance_breakdown}).
    \item Additional qualitative examples on the Epic-Kitchens and the Something-Something datasets (\S\ref{supp:qualitative_examples}).
    % \item Implementation details, extended from Section~4 in the main paper (\S\ref{supp:implementation}).
\end{itemize}

\section{Datasets}\label{supp:datasets}

\subsection{Epic-Kitchens}
EPIC-Kitchens is a large-scale benchmark dataset consisting of 700 videos recorded by 32 participants \cite{damen2018scaling, damen2020rescaling}, totalling 100 hours of egocentric videos capturing daily activities in kitchen environments. In our experiments, we use the annotations for the action recognition task. It features compositional actions which can be broken down into the noun (the active object participating in the action, e.g. ``carrot'', ``pan'', etc.) and the verb (the activity itself, e.g. ``cutting'', ``washing''). In total, there are 300 noun and 97 verb categories, while the training and validation set contain $\sim$~68k and $\sim$~10k videos respectively. The modalities we use in our experiments include the RGB frames, the audio, and the optical flow. We extract the audio directly from the mp4 files, and use the optical flow as released by the dataset authors \cite{damen2020rescaling}.

\subsubsection{Epic-Kitchens Unseen Participants}

This particular subset of the Epic-Kitchens validation split contains 1065 action sequences from two participants which were not observed in the training dataset (i.e. videos recorded by them are not included in the training data). We use this data split to more explicitly gauge the compositional generalization performance of the models. Namely, standard RGB models tend to pick up undesirable biases to discriminate between different actions, i.e. objects or environment cues unrelated to the action \cite{materzynska2020something}. Using the Epic-Kitchens Unseen Participants split, we verify the extent to which students distilled from multimodal teachers are robust w.r.t. this type of distribution shift.

\subsection{Something-Something}
The Something-Something V2 \cite{goyal2017something} dataset consists of (mainly) egocentric videos of people performing 174 unique object-agnostic actions with their hands, e.g. ``pushing [something] left'', ``taking [something] out of [something]''. Notice that the action classes do not account for specific objects, but rather, only for the activity itself. Therefore, on Something-Something, there is increased focus on capturing temporal relationships that characterize the actions. The training and validation set contain $\sim$~169k and $\sim$~26k videos respectively. Furthermore, to deal with the environment bias (models relying on unrelated environmental cues to discriminate between the actions), videos recorded by the same participant can be in either the training or validation set. Nevertheless, the objects the participants interact with -- even though unrelated to the action label -- can appear in both the training and the testing data, indicating that models that observe the videos through the RGB modality can overfit on the objects' appearance.

\subsection{Something-Else}
In Something-Something, the objects present in the scene (where the action takes place) may appear in both the training and the testing data. The goal of Something-Else \cite{materzynska2020something} is to deal with the issue of models exploiting visual cues related to objects' appearance. Materzynska \etal \cite{materzynska2020something}, propose a data split according to the objects’ distribution at training and test time. The data is divided such that the models encounter distinct objects during training and testing. The training and validation set contain $\sim$~55k and $\sim$~58k videos respectively, with 174 action categories. This data split is explicitly aimed at testing the compositional generalization of the models. Furthermore, Materzynska \etal \cite{materzynska2020something}, show that a standard RGB-based model \cite{carreira2017quo} exhibits significantly lower performance on the Something-Else split compared to the standard Something-Something split. To improve the generalization ability of standard RGB-based models, the work of Materzynska \etal \cite{materzynska2020something}, as well as subsequent works \cite{radevski2021revisiting, herzig2022object}, propose using object detections as input to the model \cite{ren2015faster}, as they are agnostic to the appearance of individual objects.
\section{Data Modalities \& Models}\label{supp:modalities}

\subsection{RGB frames (RGB)}
To encode the RGB frames we follow the standard setting of \cite{liu2021swin} for all datasets -- Epic-Kitchens (including the Unseen split), Something-Something (including the Something-Else compositional generalization split). Unless stated otherwise, we use the Swin-T \cite{liu2021swin} model to process the RGB frames. During training, we resize the image such that the shorter dimension (typically the height) is set to a value randomly chosen from the interval $[224, 320]$, and subsequently select a random $224 \times 224$ crop. Additionally, we adopt random horizontal flips with probability $50$\% (only for Epic-Kitchens), and color jittering. During inference, we resize the image such that the shorter dimension (typically the height) is set to $224$, and then select a $224 \times 224$ central crop for each frame.\par
In the case of the R3D \cite{kataoka2020would} models, where we focus on testing our approach on computationally cheaper and faster architectures and settings, we keep the same train and inference setup, with the exception of the final crop size which we reduce to $112$, as per Kataoka \etal \cite{kataoka2020would}.

\subsection{Optical Flow (OF)}
We process the optical flow frames in the same fashion as the RGB frames and use the same vision backbone (Swin-T) for both Epic-Kitchens (including the Unseen split) and Something-Something (including the Something-Else compositional generalization split). We use the two components of the velocity as the first two channels of the input, and in order to maintain the same architecture, we append an additional channel where we set each pixel intensity to 0.0, effectively expanding the number of input channels to 3. We use the same data augmentations as with the RGB model, with the exception of color jitter. %We always use the Swin-T model as the backbone of our optical flow model.

\subsection{Audio (A)}
On Epic-Kitchens, we first convert the $24000$Hz stereo audio to $16000$Hz monoaural audio. We compute the mel-spectrograms of audio segments using 1024 FFT bins and 128 mel filter banks. We use the Hann window with length of 160, with an 80 sample overlap between successive windows. We square the magnitude after computing the FFT, and thus obtain the signal power at each frequency bin for each time-step. For the audio segments of $1.116$ with the sample frequency of $16000$, we thus obtain spectrograms with $128$ frequency bins and $224$ timesteps.

During training, as data augmentation, we perform random time and frequency masking of the spectrograms, as per the work of \cite{park2019specaugment}. In time masking, with a probability of $50$\%, we randomly chose the number of masked time-steps $T_n$ from the range $[ 0, 80 ]$, and the starting time-step from the range $[ 0, 224 - T_n )$, such that, for all the frequency bins, the range of time-steps $[ T_s, T_s + T_n )$ is masked by setting the power value in the spectrogram to 0. In frequency masking, with a probability of $50$\%, we randomly chose the number of masked frequency bins $F_n$ from the range $[0, 80]$, and the starting frequency bin $F_s$ from the range of $[ 0, 128 - F_n )$, such that, for all the timesteps, the range of bins $[ F_s, F_s + F_n )$ is masked by setting the power value in the spectrogram to 0. Afterwards, we resize the spectrogram height to a value randomly chosen from the interval $[224, 320]$, and finally select a random $224 \times 224$ crop. 

During inference, we do not perform time and frequency masking, we simply resize the height of the spectrogram to $224$ and select a $224 \times 224$ central crop for each frame.

We use the obtained spectrogram repeated 3 times to construct a 3-channel input for the Swin-T backbone. Despite the simple setup, our audio-specific model performs on-par with more sophisticated state-of-the-art audio models \cite{stergiou2022play} on Epic-Kitchens.\par

\subsection{Object Detections (OBJ)}
When pre-processing the Object Detections (on Something-Something and Something-Else) we closely follow the setup of \cite{radevski2021revisiting}. We represent each video frame with only its object detections -- bounding boxes \& object categories. We use the object detections released from Herzig \etal \cite{herzig2022object} for Something-Something and Materzynska \etal \cite{materzynska2020something} for Something-Else, which had been obtained using a Faster R-CNN \cite{ren2015faster}, trained as per the setting of \cite{shan2020understanding}. We use the STLT (Spatial-Temporal Layout Transformer) model to encode the object detections, while following the settings and the implementation of \cite{radevski2021revisiting}.\par
In the STLT model, one Transformer model \cite{vaswani2017attention} encodes the spatial relations between the objects in each frame independently, while another Transformer encodes the temporal relations given the embedding of each frame (output of the Spatial-Transformer).

\section{Loss Term Weighting - Something-Something \& Something-Else}\label{supp:weighting}
In the vein of the ablation study reported in Table~5 (main paper), we conduct experiments on the Something-Something and the Something-Else datasets. Namely, the main findings from Table~5 suggest that (i) training with the ground-truth labels cross-entropy loss, in conjunction with the multimodal knowledge distillation loss, overcomes the issue of inferior modality-specific teachers, and (ii) weighting the teachers in the ensemble (such that their each individual cross-entropy loss on a holdout set of $Z = 1000$ samples are minimized) improves the performance further. 
% In Table~\ref{table_supp:ablation-weighting}, we observe that for Something-Something and Something-Else -- where each model in the teacher is ensemble is complementary to the others -- the ground truth loss to have little to no effect on the performance. 
In Table~\ref{table_supp:ablation-weighting}, however, we observe that in the case of Something-Something and Something-Else, the addition of the loss term featuring ground-truth labels has a small effect on the performance of the student. Namely, as discussed in the main paper, on both the Something-Something and the Something-Else datasets, all modality-specific models perform well, and are complementary to each other, therefore, there is a lesser need for joint training using the ground truth labels and the distillation loss.

\begin{table}[t]
\centering
\resizebox{0.85\columnwidth}{!}{
\begin{tabular}{>{\columncolor{graygray}}c>{\columncolor{pinkpink}}c>{\columncolor{yellowyellow}}c>{\columncolor{blueblue}}c>{\columncolor{blueblue}}c>{\columncolor{blueblue}}c>{\columncolor{blueblue}}c} \toprule
    \cellcolor{white}{} & \cellcolor{white}{} & \cellcolor{white}{} & \multicolumn{2}{c}{Something-Something} & \multicolumn{2}{c}{Something-Else} \\ \midrule
    {Objective} & {$\lambda$} & {$\gamma$} & {Action@1} & {Action@5} & {Action@1} & {Action@5} \\ \midrule
    $\mathcal{L}_{\text{CE}}$ & 0.0 & \NA & 60.3 & 86.4 & 51.8 & 79.5 \\
    $\mathcal{L}_{\text{KL}}$ & 1.0 & 30.0 & 63.0 & 88.9 & 59.1 & 86.1 \\
    $\mathcal{L}_{\text{CE}} \wedge \mathcal{L}_{\text{KL}}$ & 0.8 & 30.0 & 63.1 & 88.3 & 59.3 & 86.3 \\ \bottomrule
    % $\mathcal{L}_{\text{CE}} \wedge \mathcal{L}_{\text{KL}}$ & 0.8 & 1.0 & X & X & X & X \\ \bottomrule
\end{tabular}
}
\caption{Ablation study on Something-Something and Something-Else; $\lambda$: Distillation and Cross-Entropy loss balancing term; $\gamma$: Temperature of the Ensemble Teacher Weighting.}
\label{table_supp:ablation-weighting}
\end{table}
\section{Details on action recognition models trained on multimodal data}\label{supp:omnivore}
Multiple works explore a similar setting, i.e. using multiple modalities for training while performing inference using only RGB frames. Some of the most prominent works are ModDrop \cite{neverova2015moddrop}, DMCL \cite{garcia2018modality, garcia2019dmcl}, and Omnivore \cite{girdhar2022omnivore}.\par
\textbf{ModDrop.} Neverova \etal \cite{neverova2015moddrop} propose a method where a multimodal model is made robust to missing modalities during inference by randomly dropping out modalities during training. Namely, the model is trained such that it might observe all modalities, a partial set of modalities or only a single modality during training. This makes the model recognize cues, generally multimodal, from RGB data, and is therefore superior to an RGB model.\par
\textbf{DMCL.} Garcia \etal \cite{garcia2018modality, garcia2019dmcl} propose a four-step mulitmodal distillation framework which is tested on non-egocentric data. They train a model on multimodal inputs, where for each training video-action sample, the teacher network is established as the model which exhibits the lowest cross-entropy w.r.t. ground truth action, and the remaining models are the students. Then, the student models are trained on the soft teacher labels. On the other hand, our method is simple -- standard knowledge distillation -- and flexible -- other models can easily be added to the ensemble and the student model can be retrained while keeping the existing models fixed.\par
\textbf{Omnivore.} \cite{girdhar2022omnivore} To the best of our knowledge, Omnivore is the latest and best performing method that uses multimodal data during training, while using only unimodal data during inference. Compared to multimodal distillation, Omnivore can perform inference using a single set of weights across all different modalities it was trained on. In particular, Girdhar \etal \cite{girdhar2022omnivore} use multimodal data during pre-training, while for the downstream task, the model is directly fine-tuned on the RGB frames. The resulting model -- pre-trained on omnivorous data -- is superior. In our work, to establish an Omnivore baseline parallel to multimodal distillation, we perform training on the multimodal data the downstream task features.\par
To train a single model (single set of weights) using multimodal data, Girdhar \etal \cite{girdhar2022omnivore} propose two strategies to sample the batches: (i) Batches contain data of mixed modalities -- heterogenous batches, or (ii) each batch is unimodal -- homogenous -- with a randomly chosen modality. In our work, we found that (i) yields a model with performance similar to the simply training the model on RGB frames, and therefore, we opted for (ii).
\section{Learning curves}\label{supp:learning_curves}
\begin{figure*}[t]
\centering
\includegraphics[width=1.0\columnwidth]{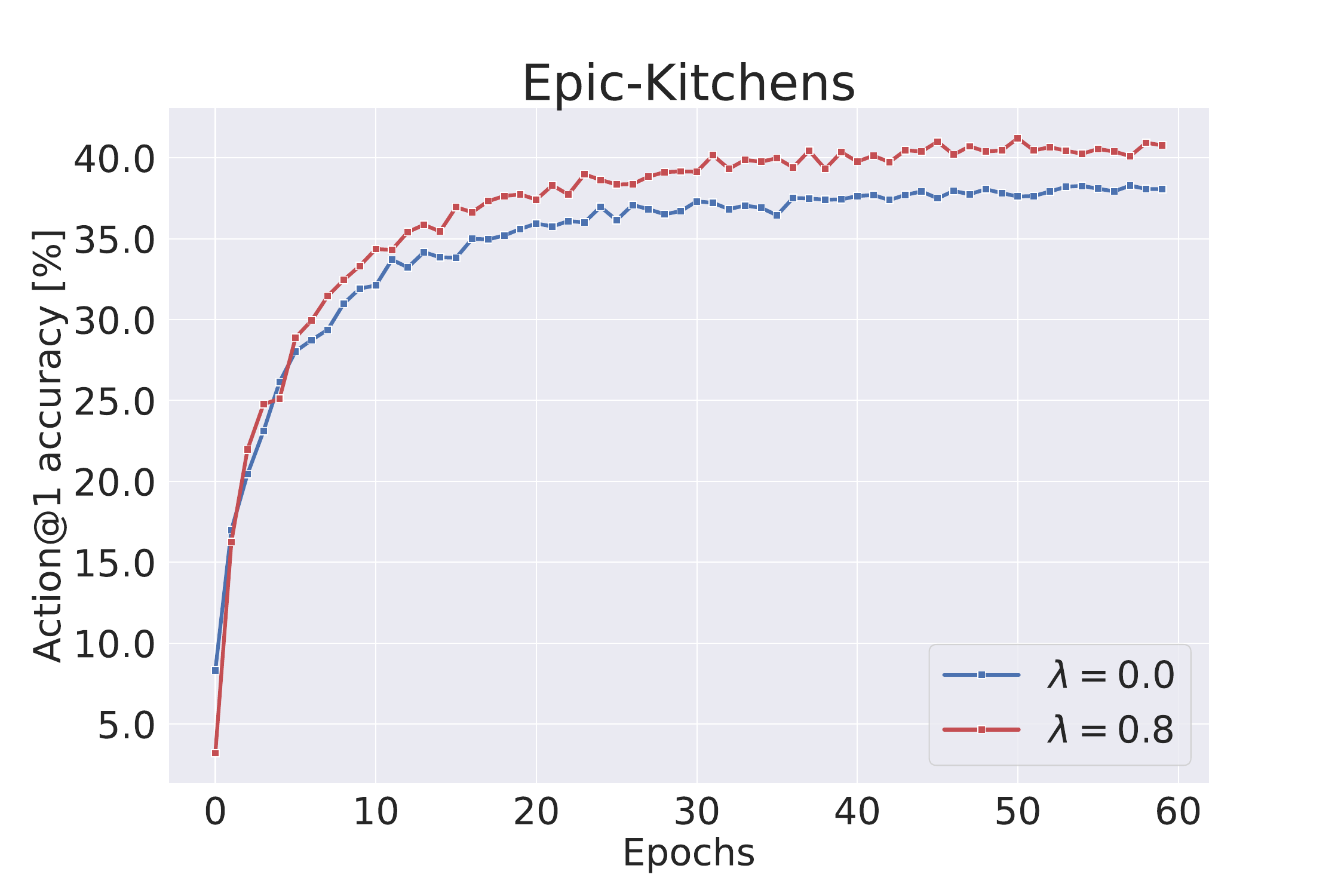}
\includegraphics[width=1.0\columnwidth]{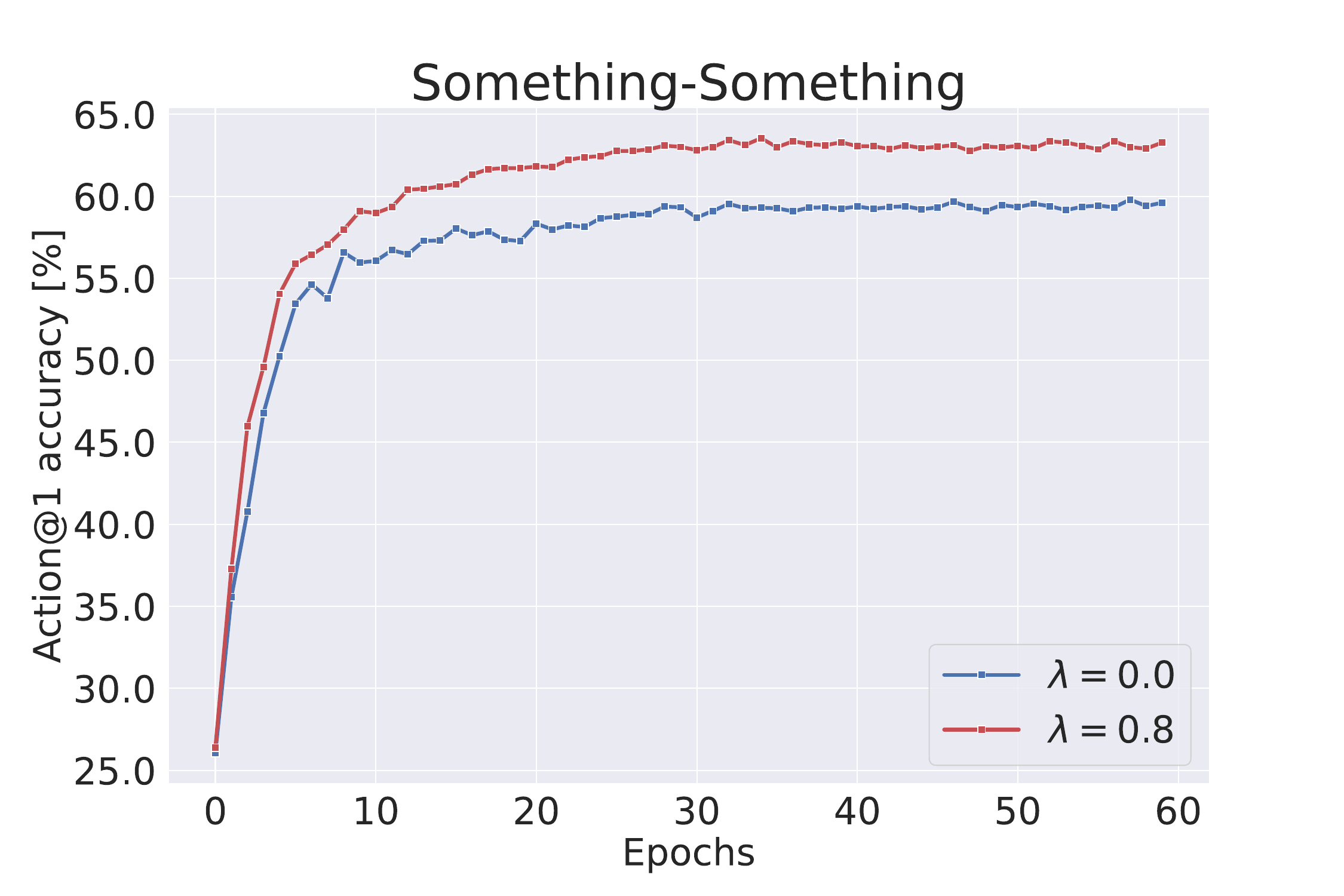}
\caption{Learning curve on the Epic-Kitchens dataset (left) and the Something-Something dataset (right).}
\label{fig:learning-curves}
\end{figure*}

To ensure better reproducibility, we also report learning curves on the Epic-Kitchens and Something-Something dataset. That is, in Figure~\ref{fig:learning-curves} we report the top-1 validation set accuracy measured at the end of each epoch on the y-axis, and the number of executed training epochs on the x-axis. We observe that for both datasets, the distilled student converges to a model which generalizes better than training on the ground truth labels alone.

\section{Per-Class Performance Breakdown}\label{supp:performance_breakdown}

\begin{figure*}[t]
\centering
\includegraphics[width=1.0\textwidth]{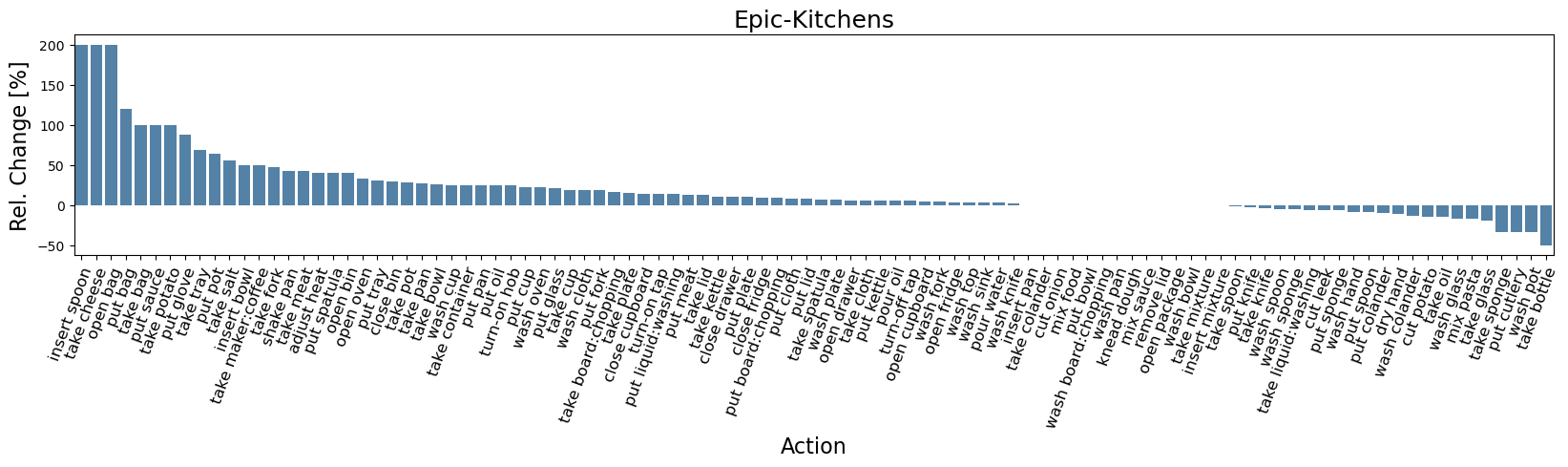}
\caption{Per-class performance change between the student and the RGB baseline on the Epic-Kitchens regular split.}
\label{fig:per_class_epic_regular}
\end{figure*}

\begin{figure*}[t]
\centering
\includegraphics[width=1.0\textwidth]{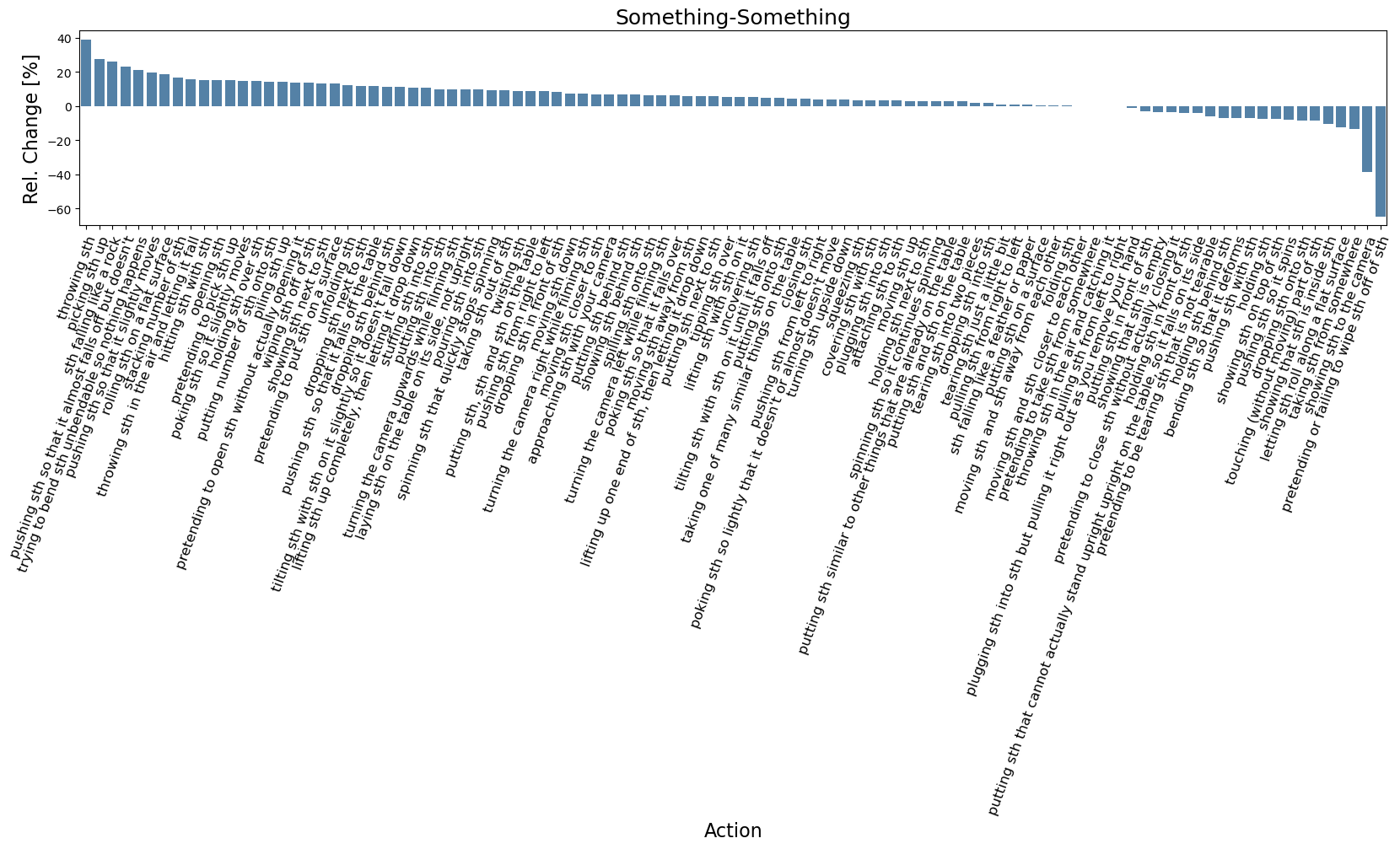}
\caption{Per-class performance change between the student and the RGB baseline on the Something-Something dataset.}
\label{fig:per_class_sth_sth}
\end{figure*}

In Figure~4 in the main paper, we provided a per-class performance breakdown of the 20 most frequent actions for Epic-Kitchens and Something-Something. Here, in Figure~\ref{fig:per_class_epic_regular} and Figure~\ref{fig:per_class_sth_sth}, we provide an extended per-class performance breakdown of the 100 most frequent actions. On both datasets, we observe that the student is superior to the model trained on the ground truth labels on the majority of action classes.

\section{Qualitative Examples}\label{supp:qualitative_examples}

\begin{figure*}[t]
\centering
\includegraphics[width=1.0\textwidth]{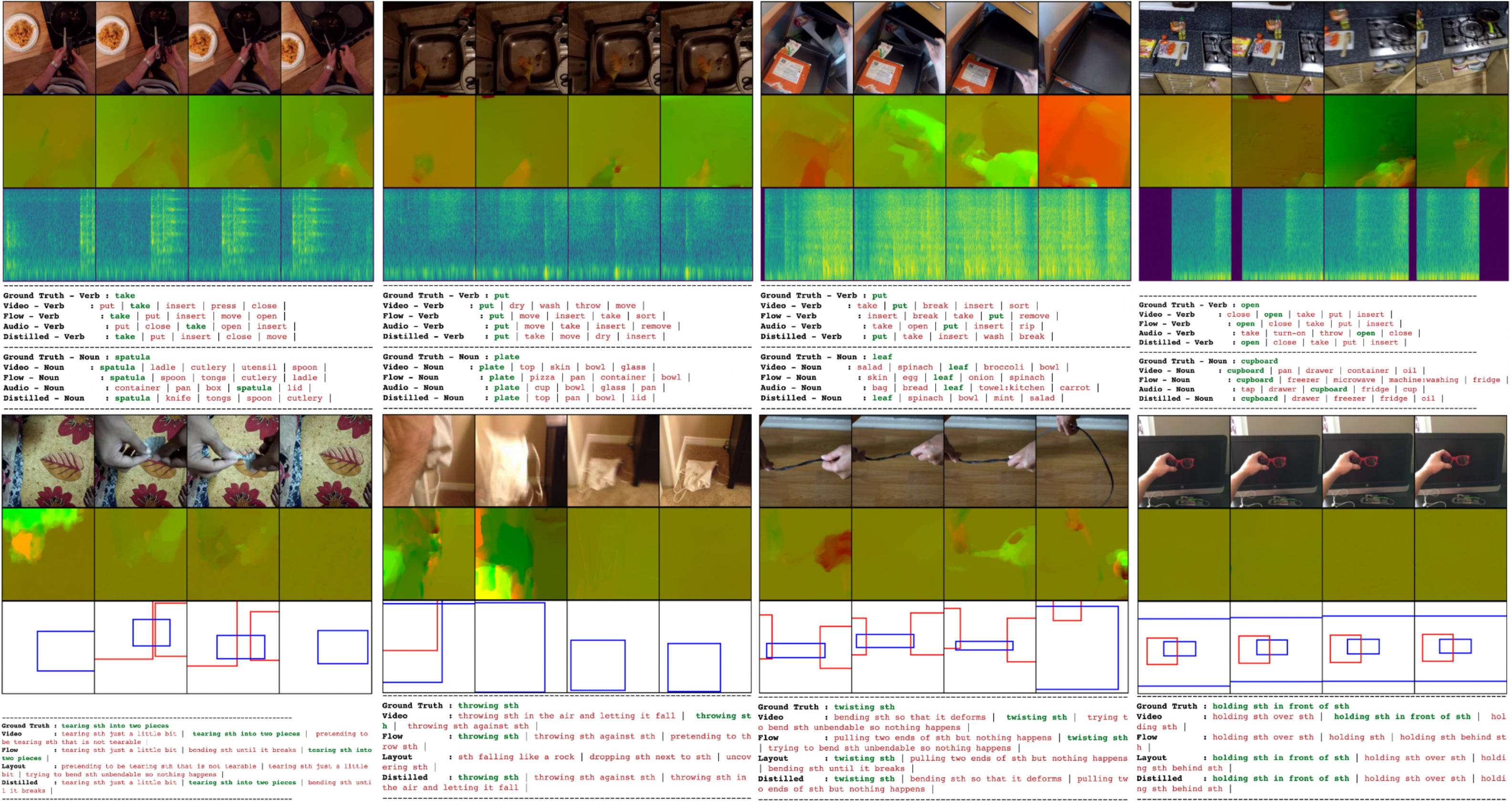}
\caption{Qualitative evaluation for Epic-Kitchens (top) and Something-Something (bottom).}
\label{fig:qualitative-supplementary}
\end{figure*}

We report additional qualitative examples in Figure~\ref{fig:qualitative-supplementary} supplementing the results of Figure~7 in the main paper.

% Notice the interesting scenario of the right-most Something-Something example, where we observe that the flow model is unable to accurately predict the action category due to there being no movement in the scene. Further, the layout model is able to accurately predict the action (cued through the static scene-layout). Finally, the distilled student can accurately predict the action.

\end{document}